\documentclass[letterpaper]{article} 
\usepackage{aaai25}  
\usepackage{times}  
\usepackage{helvet}  
\usepackage{courier}  
\usepackage[hyphens]{url}  
\usepackage{graphicx} 
\usepackage{natbib}  
\usepackage{caption} 
\usepackage{algorithm}
\usepackage{algorithmic}

\usepackage{newfloat}
\usepackage{listings}
\DeclareCaptionStyle{ruled}{labelfont=normalfont,labelsep=colon,strut=off} 
\floatstyle{ruled}
\newfloat{listing}{tb}{lst}{}
\floatname{listing}{Listing}
\usepackage{subfigure}
\usepackage{amsmath}
\usepackage{amssymb}
\usepackage{booktabs}
\usepackage{bm}

\title{Federated Unlearning with Gradient Descent and Conflict Mitigation}
\author {
Zibin Pan\textsuperscript{\rm 1,2},
Zhichao Wang\textsuperscript{\rm 1},
Chi Li\textsuperscript{\rm 1,6},
Kaiyan Zheng\textsuperscript{\rm 5},
Boqi Wang\textsuperscript{\rm 1},\\
Xiaoying Tang\thanks{Xiaoying Tang and Junhua Zhao are corresponding authors.}\textsuperscript{\rm 1,3,4},
Junhua Zhao$^*$\textsuperscript{\rm 1,3}
}
\affiliations {
\textsuperscript{\rm 1}The School of Science and Engineering, The Chinese University of Hong Kong, Shenzhen, China\\
\textsuperscript{\rm 2}The Cyberspace Academy, Guangzhou University\\
\textsuperscript{\rm 3}The Shenzhen Institute of Artificial Intelligence and Robotics for Society\\
\textsuperscript{\rm 4}The Guangdong Provincial Key Laboratory of Future Networks of Intelligence\\
\textsuperscript{\rm 5}Department of Mathematics and Department of Statistics, University of Michigan\\
\textsuperscript{\rm 6}Shenzhen Research Institute of Big Data\\
zibinpan@link.cuhk.edu.cn, zhichaowang@link.cuhk.edu.cn, chili@link.cuhk.edu.cn, kaiyanz@umich.edu, boqiwang@link.cuhk.edu.cn, tangxiaoying@cuhk.edu.cn, zhaojunhua@cuhk.edu.cn
}

\usepackage{bibentry}

\begin{document}

\maketitle

\begin{abstract}
	Federated Learning (FL) has received much attention in recent years. However, although clients are not required to share their data in FL, the global model itself can implicitly remember clients' local data. Therefore, it's necessary to effectively remove the target client's data from the FL global model to ease the risk of privacy leakage and implement ``the right to be forgotten". Federated Unlearning (FU) has been considered a promising way to remove data without full retraining. But the model utility easily suffers significant reduction during unlearning due to the gradient conflicts. Furthermore, when conducting the post-training to recover the model utility, the model is prone to move back and revert what has already been unlearned. To address these issues, we propose Federated Unlearning with Orthogonal Steepest Descent (FedOSD). We first design an unlearning Cross-Entropy loss to overcome the convergence issue of the gradient ascent. A steepest descent direction for unlearning is then calculated in the condition of being non-conflicting with other clients' gradients and closest to the target client's gradient. This benefits to efficiently unlearn and mitigate the model utility reduction. After unlearning, we recover the model utility by maintaining the achievement of unlearning. Finally, extensive experiments in several FL scenarios verify that FedOSD outperforms the SOTA FU algorithms in terms of unlearning and model utility.
\end{abstract}

%
\begin{links}
	\link{Code}{https://github.com/zibinpan/FedOSD}
\end{links}

\section{Introduction}
Federated Learning (FL) has increasingly gained popularity as a machine learning paradigm in recent years \cite{FedAvg}. It allows clients to cooperatively train a global model without sharing their local data, which helps address data island and privacy issues \cite{yu2022survey}. But previous studies demonstrate that clients' local training data is inherently embedded in the parameter distribution of the models trained on it \cite{de2021impact,MoDe}. Therefore, in light of privacy, security, and legislation issues, it's necessary to remove clients' training data from the trained model \cite{FedRecovery}, especially when clients opt to withdraw from FL. This is known as the right to be forgotten (RTBF) \cite{FedEraser}, which is enacted by privacy regulations such as the General Data Protection Regulation (GDPR) \cite{GDPR} and the California Consumer Privacy Act (CCPA) \cite{CCPA}.

A naive way to achieve this goal is to retrain the FL model. But it brings large computation and communication costs \cite{FUsurvery2}. In contrast, unlearning is a more efficient way, which has been well studied in centralized machine learning \cite{bourtoule2021machine}. Inspired by it, Federated Unlearning (FU) has emerged, aiming to remove data from a trained FL model while trying to maintain model utility.
\begin{figure}[t!]
	\centering
	\includegraphics[width=0.75\columnwidth]{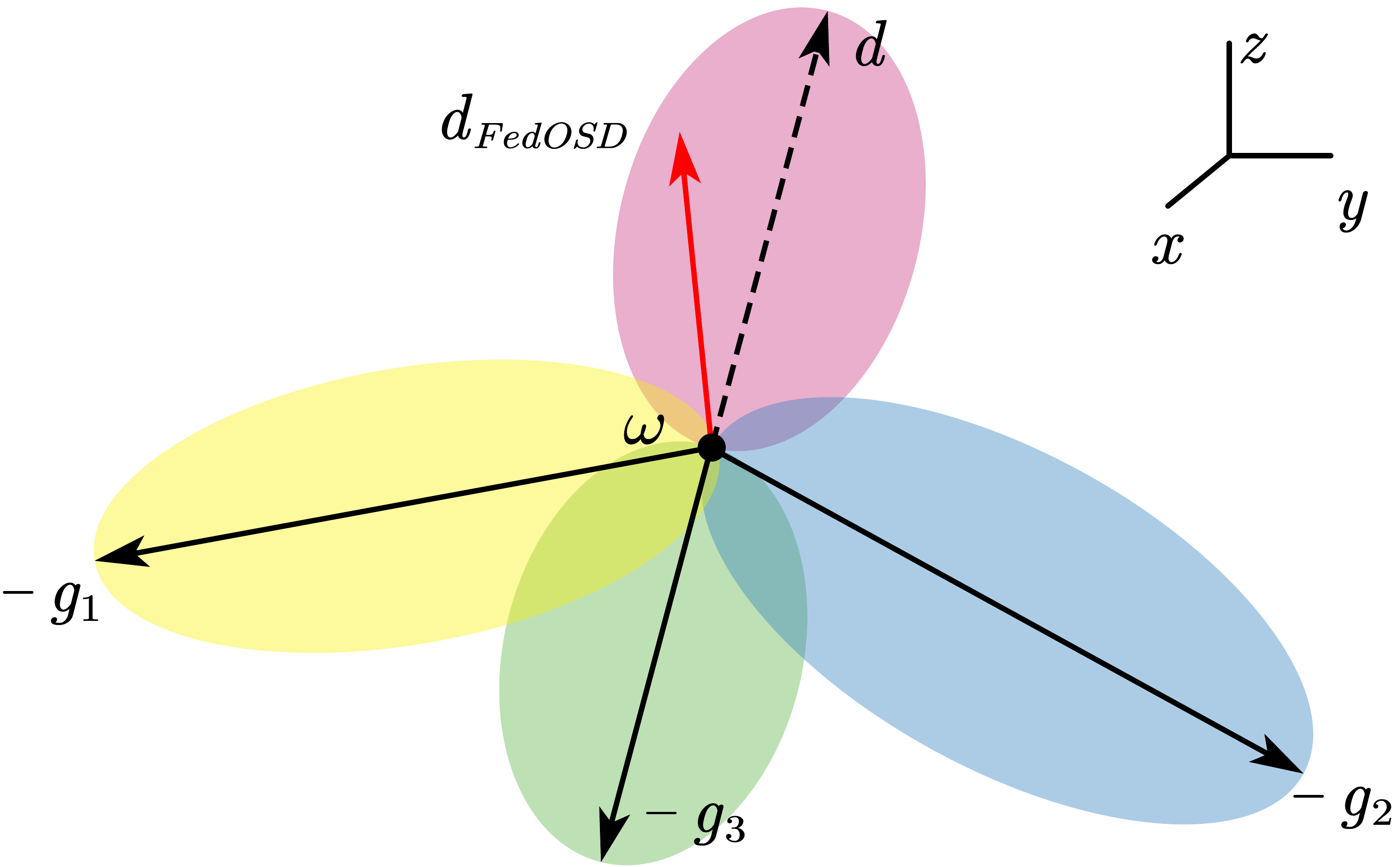}
	\caption{A demo of three clients. $g_1, g_2, g_3$ represent the gradient of clients. $d$ denotes the update direction for unlearning client $3$, which is conflicting with $g_1$ and $g_2$, i.e., $g_1\cdot d < 0$ and $g_2\cdot d < 0$. $d_{FedOSD}$ represents the direction obtained by FedOSD, which doesn't conflict with $g_1$ and $g_2$.}
	\label{fig:unlearning_direction}
\end{figure}

In this context, numerous FU techniques have been proposed. Federaser \cite{FedEraser} leverages the norms of historical local updates in the previous FL training to accelerate retraining. FedKdu \cite{FedKdu} and FedRecovery \cite{FedRecovery} utilize the historical gradients to calibrate the model to erase the training data of the target client (i.e., the client that requests for unlearning). However, these methods require clients to continuously record historical information during FL training \cite{FUsurvey1}. Moreover, \cite{MoDe} propose MoDe to unlearn the target client's data by momentum degradation, but it requires simultaneously retraining a model for updating the unlearning model, which brings additional communication costs.

Among prior studies, Gradient Ascent (GA) is considered a viable and efficient method for FU \cite{FUsurvery2}, which formulates unlearning as the inverse process of learning and takes the inverse of the loss function to reduce the model performance on the target client. It can effectively achieve the unlearning goal in few communication rounds while not bringing extra storage costs \cite{FUPGA}. However, we observe that there exist the following three primary challenges when performing GA in FU.

\textbf{Challenge 1: Gradient explosion.}
Gradient explosion is a significant challenge for GA-based federated unlearning, necessitating a substantial reliance on experimental hyper-parameter tuning. This is because the loss function generally has no upper bound (see Fig.~\ref{fig:ce_loss}(a)). Consequently, executing GA to unlearn results in gradient explosion and cannot converge. We delve further into this in Section \ref{sec:UCE_loss}. To this end, \cite{FUPGA} project model parameters to an $L_2$-norm ball of radius $\delta$. But it requires experimentally tuning $\delta$.

\textbf{Challenge 2: Model utility degradation.}
Directly applying GA to unlearn would inevitably destroy the model utility \cite{FUsurvey1}, even leading to catastrophic forgetting \cite{FUsurvery2}. Specifically, the model performance on remaining clients (i.e., those that do not require unlearning) would decrease heavily. One direct cause is the gradient conflict \cite{FedMDFG}, where the model update direction for unlearning a client conflicts with those of the remaining clients, directly leading to a reduction in model utility. Fig.~\ref{fig:unlearning_direction} illustrates an example in which client 3 requests unlearning, but the model update direction conflicts with the gradients of client 1 and client 2. Consequently, the updated model would exhibit diminished performance on client 1 and 2.

\textbf{Challenge 3: Model reverting issue in post-training.}
After unlearning, post-training is often conducted, where the target client leaves and the remaining clients continually train the FL global model cooperatively to recover the model utility that was reduced in the previous unlearning \cite{FUPGA,FedKdu}. However, we observe that during this stage, the model tends to revert to its original state, resulting in the recovery of previously forgotten information and thus losing the achievement of unlearning. This issue is further explored in Section \ref{sec:post-training}.

To handle the aforementioned challenges, we propose the \textbf{Fed}erated Unlearning with \textbf{O}rthogonal \textbf{S}teepest \textbf{D}escent algorithm (FedOSD). Specifically, to handle the gradient explosion inherent in GA, we modify the Cross-Entropy loss to an unlearning version and employ the gradient descent, rather than GA, to achieve the unlearning goal. Subsequently, an orthogonal steepest descent direction that avoids conflicts with retained clients' gradients is calculated to better unlearn the target client while mitigating the model utility reduction. In post-training, we introduce a gradient projection strategy to prevent the model from reverting to its original state, thereby enabling the recovery of model utility without compromising the unlearning achievement.

Our contributions are summarized as follows:
\begin{enumerate}
	\item We introduce an Unlearning Cross-Entropy loss that can overcome the convergence issue of Gradient Ascent.
	\item We propose FedOSD that establishes an orthogonal steepest descent direction to accelerate the unlearning process while mitigating the model utility reduction.
	\item We design a gradient projection strategy in the post-training stage to prevent the model from reverting to its original state for better recovering the model utility.
	\item We implement extensive experiments on multiple FL scenarios, validating that FedOSD outperforms the SOTA FL unlearning approaches in both unlearning performance and the model utility.
\end{enumerate}

\section{Background \& Related Work}
\subsection{Federated Learning (FL)}
The traditional FL trains a global model $\omega$ cooperatively by $m$ clients, which aims to minimize the weighted average of their local objectives \cite{li2020federated}: $\min\nolimits_\omega \sum\nolimits_{i = 1}^m p_i L_i(\omega)$,
where $p_i$$\geq$$0$, $ \sum\nolimits_{i=1}^{m} p_i$$=$$1$. $L_i$ is the local objective of client $i$, which is usually defined by the empirical risks over the local training data with $N_i$ samples: $L_i(\omega^t)$$=$$\sum\nolimits_{j=1}^{N_i} \frac{1}{N_i}{L_{i_j}(\omega^t)}$. $L_{i_j}$ is the loss on sample $j$, which is obtained by a specific loss function such as Cross-Entropy (CE) loss:
\begin{equation}
	\label{equ:CE_Loss}
	L_{CE} = -\sum\nolimits_{c = 1}^{C} y_{o,c} \cdot log(p_{o,c}),
\end{equation}
where $C$ denotes the number of classes. $y_{o,c}$ is the binary indicator (0 or 1) if class label $c$ is the correct classification for observation $o$, i.e., the element of the one-hot encoding of sample $j$'s label. $p_{o,c}$ represents the predicted probability observation $o$ that is of class $c$, which is the $o^{th}$ element of the softmax result of the model output.

\subsection{Federated Unlearning} \label{sec:Federated Unlearning}
Federated Unlearning (FU) aims to erase the target training data learned by the FL global model, while mitigating the negative impact on the model performance (e.g., accuracy or local objective). Recognized as a promising way to protect `the Right to be Forgotten' of clients, FU can also counteract the impact of data poisoning attacks to enhance the security \cite{FUsurvey1,FUsurvery2}.

FU has garnered increasing interest in recent years. (1) Some previous studies have leveraged the historical information of FL training to ease the target client's training data, such as FedEraser \cite{FedEraser}, FedKdu \cite{FedKdu}, FedRecovery \cite{FedRecovery}, etc. (2) Besides, \cite{MoDe} adopt momentum degradation to FU. (3) \cite{su2023asynchronous} use clustering and (4) \cite{ye2024heterogeneous} employ distillation to unlearn. (5) A significant approach related to our work is Gradient Ascent, which utilizes the target client's gradients for unlearning \cite{FUPGA,EWCSGA}.

Based on the types of client data that need to be forgotten, Federated Unlearning can be categorized into sample unlearning and client unlearning \cite{FUsurvery2}. We focus on client unlearning in this paper for two reasons. First, we can make a fair comparison with previous record-based FU methods such as FedEraser and FedRecovery. Since they rely on pre-recording information like model gradients on the target data that needed to be unlearned, they are not suitable for sample unlearning. Since in the sample unlearning, clients only request to unlearn partial training data. However, no one knows which data will be requested to unlearn during FL training, and thus preparing these records in advance for later unlearning is not feasible in practice.

Furthermore, for other FU algorithms that do not necessitate using historical training records, we can technically treat unlearning samples as belonging to a virtual client. Hence, the sample unlearning can be transferred to the client unlearning. For example, when a client requests to unlearn partial data $D_{u}$, we can form a new virtual client $u$ that owns $D_{u}$ and unlearn it. 

The formulation of unlearning the target client $u$ from the trained global model can be defined by:
\begin{equation}
	\label{Problem:unlearning_objective}
	\max\limits_\omega~ L_u(\omega),
\end{equation}
where $L_u(\omega)$ represents the local objective of client $u$ in FL.

\begin{figure}[t!]
	\centering
	\includegraphics[width=1.0\columnwidth]{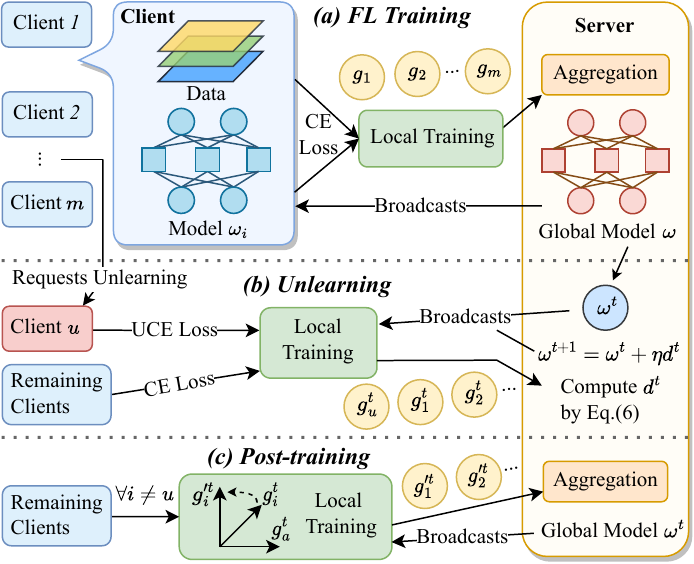}
	\caption{The FedOSD framework comprises two main stages: (b) the unlearning stage and (c) the post-training stage. Subfigure (a) depicts the previous FL training procedure before the client requests for unlearning, where the obtained model is denoted as $\omega^0$ and serves as the original model for unlearning.}
	\label{fig:framework}
\end{figure}

\textbf{Federated Unlearning with Gradient Ascent.}
Gradient Ascent (GA) \cite{EWCSGA,FUsurvery2} is a proactive and efficient approach for solving Problem (\ref{Problem:unlearning_objective}). At each communication round $t$, it strives to maximize the empirical loss of the target client $u$ by updating the model according to $\omega^{t+1} = \omega^t + \eta^t \nabla L_u(\omega^t)$ with the step size (learning rate) $\eta^t$ \cite{FUPGA}. However, \cite{EWCSGA} suggest that this approach would fail because of destroying the global model performance for the remaining clients. To this end, they propose EWCSGA, which incorporates a regularization term to the cross entropy loss to mitigate the negative impact on the model utility. Besides, another approach computes an update direction $\Delta \omega$ orthogonal to the subspace of the model layer inputs $x$, i.e., $\Delta \omega x=0$ \cite{saha2021gradient,SFU}. This kind of method works well in protecting the model utility in centralized learning, however, it is not suitable for FL due to potential privacy leakage from uploading $x$. SFU \cite{SFU} attempts to mitigate this issue by multiplying $x$ with a factor $\lambda$ before uploading, but attackers can easily recover the original data. Additionally, it would suffer model utility reduction during unlearning, because the derived model update direction is only orthogonal to a subset of the input data from the remaining clients, which cannot ensure the preservation of model utility. We validate these points through the experimental results presented in Table \ref{tab:sfu}.

Our method draws from the idea of GA to achieve the goal of unlearning. Differently, we modify the CE loss function to an unlearning version to overcome the gradient explosion issue, and compute the steepest descent direction that not only aligns closely with the target client's gradient but also avoids conflicts with the retained clients' gradients. This approach enables more effective unlearning while mitigating the model utility degradation.

\begin{algorithm}[t]
	\caption{\textbf{FedOSD}}
	\label{alg:algorithm}
	\begin{algorithmic}[1]
		\REQUIRE Pretrained model $\omega^0$, learning rate $\eta$, FL client set $S$, communication round $T$, max unlearning round $T_u$.
		\STATE $u \in S \leftarrow$ The client requests for unlearning.
		\FOR{$t=0,1,\cdots,T_u-1$}
		\STATE Server broadcasts $\omega^t$ to all client $i\in S$. 
		\STATE $\omega_i^t\leftarrow$ Each client $i$ performs local training, in which client $u$ switches to utilize UCE loss (Eq.~(\ref{equ:UCE_Loss})).
		\STATE \label{upload_grad} Server receives $g_i^t=(\omega^t - \omega_i^t)/\eta$ from each client $i$.
		\STATE $G \leftarrow$ concat$(g_1^t, \cdots, g_i^t, \cdots)$, $\forall i \in S, i \neq u$.
		\STATE Calculate orthogonal steepest direction $d^t$ by Eq.~(\ref{equ:osd_direction}).
		\STATE $\omega^{t+1} \leftarrow \omega^t + \eta d^t$.
		\ENDFOR
		\STATE $S \leftarrow S \backslash u$, and start the post-training stage.
		\FOR{$t=T_u, T_u+1, \cdots, T$}
		\STATE The server broadcasts $\omega^t$ to all client $i\in S$. 
		\STATE Each client $i$ performs local training to obtain $g_i^t$.
		\STATE $g_a^t\leftarrow \nabla_{\omega^t} \frac{1}{2}\|\omega^t - \omega^0\|^2$.
		\IF{$g_i^t \cdot g_a^t > 0$}
		\STATE $g_i'^t \leftarrow$ Project $g_i^t$ to the normal plane of $g_a^t$.
		\STATE Rescale $g_i'^t$ by $g_i'^t\leftarrow g_i'^t / \|g_i'^t\| \cdot \|g_i^t\|$.
		\ELSE
		\STATE $g_i'^t \leftarrow g_i^t$.
		\ENDIF
		\STATE Server receives $g_i'^t$ and aggregates $\bar g'^t = \frac{1}{|S|}\sum\nolimits_{i} g_i'^t$.
		\STATE $\omega^{t+1} \leftarrow \omega^t - \eta \bar g'^t$.
		\ENDFOR
		\ENSURE Model parameters $\omega^t$.
	\end{algorithmic}
\end{algorithm}

\section{The Proposed Approach}
Our proposed FedOSD aims to effectively remove the target client's data from the FL global model while mitigating the model performance reduction across remaining clients. Fig.~\ref{fig:framework} demonstrates the framework of FedOSD, which includes two stages: unlearning (Fig.~\ref{fig:framework}(b)) and post-training (Fig.~\ref{fig:framework}(c)). $\omega^0$ is the global model previously trained through Federated Learning across $m$ clients (Fig.~\ref{fig:framework}(a)). When client $u$ requests for unlearning, it utilizes the proposed Unlearning Cross-Entropy loss to conduct the local training. After collecting local gradients $g_i^t$, the server calculates a direction $d^t$ that is closest to client $u$'s gradient while orthogonal to remaining clients' gradients, and then updates the model by $\omega^{t+1} = \omega^t + \eta^t d^t$. In the post-training stage, a gradient projection strategy is performed to prevent the model from reverting to $\omega^0$. Detailed steps of FedOSD can be seen in Algorithm~\ref{alg:algorithm}. In Appendix.A.2, we prove the convergence of FedOSD in the unlearning and post-training stages.

\begin{figure}[t!]
	\centering
	\includegraphics[width=1.0\columnwidth]{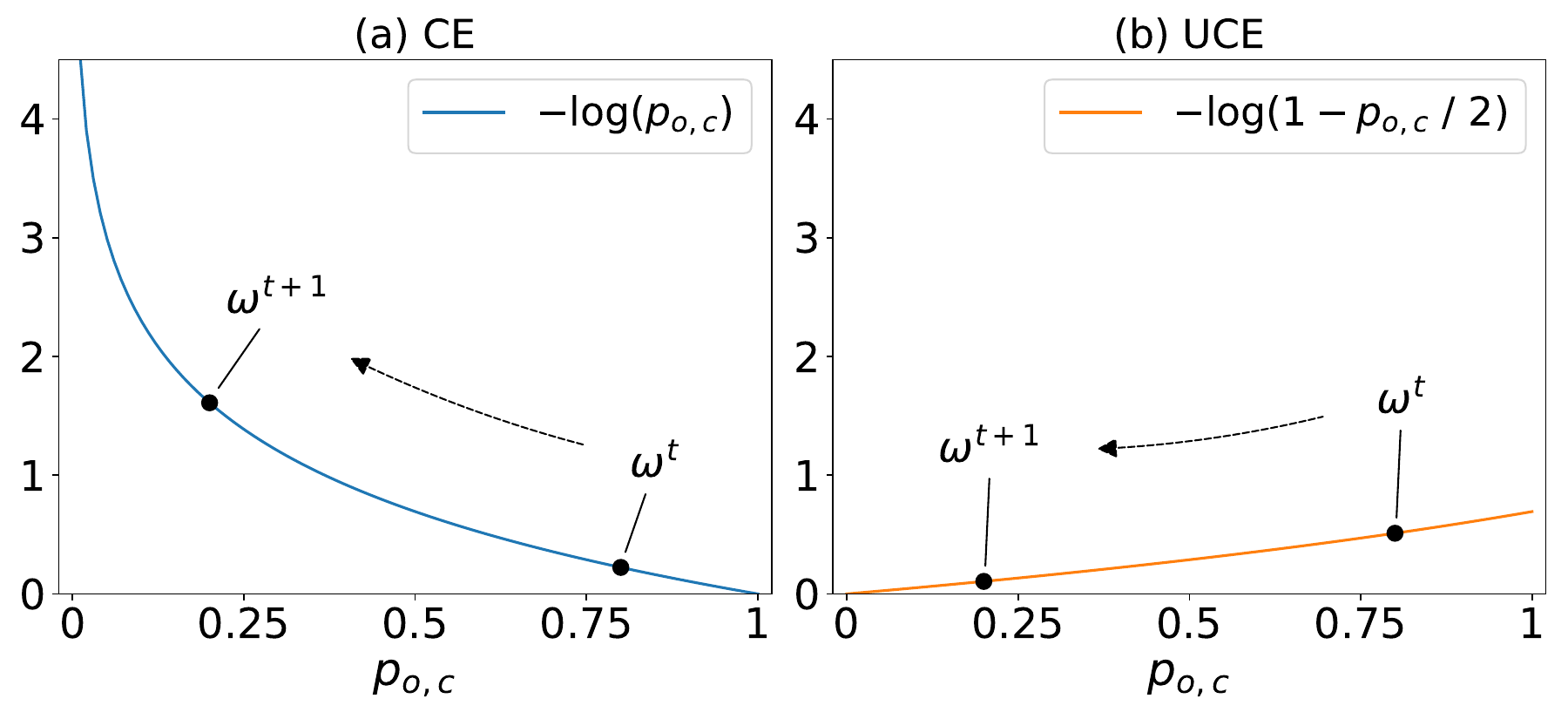}
	\caption{A comparison between (a) Cross-Entropy and (b) the proposed Unlearning Cross-Entropy. When using CE loss and GA to unlearn, it needs to drive $p_{o,c}$ to 0, leading to gradient explosion and non-convergence. When the target client switches to utilize UCE, it adopts the gradient descent to drive $p_{o,c}$ to 0 and wouldn't bring the convergence issue.}
	\label{fig:ce_loss}
\end{figure}

\subsection{Unlearning Cross-Entropy Loss} \label{sec:UCE_loss}
We first take a brief review of how Gradient Ascent can drive the model to unlearn. As shown in Fig.~\ref{fig:ce_loss}(a), by updating the global model with $\omega^{t+1} = \omega^t + \eta \nabla L_u(\omega^t)$, the local loss increases and $p_{o,c}$ approaches $0$, thus degrading the model’s prediction accuracy on the target client's data and achieving unlearning. However, the CE Loss (Eq.(\ref{equ:CE_Loss})) has no upper bound. As seen in Fig.~\ref{fig:ce_loss}(a), when $p_{o,c}$ is getting quite close to $0$, $\partial L_{CE} / \partial p_{o,c}$ would suffer the explosion and thus the local gradient of the target unlearning client explodes. That's why directly applying GA to unlearn would make the model similar to a random model \cite{FUPGA}. One conventional solution is to project the model back to an $L_2$-norm ball of radius $\delta$ \cite{FUPGA}. But it brings a hyper-parameter that requires experimentally tuning, and a fixed $\delta$ cannot guarantee the convergence.

To address this issue, we modify CE loss to an unlearning version named Unlearning Cross-Entropy (UCE) loss:
\begin{equation}
	\label{equ:UCE_Loss}
	L_{UCE} = -\sum\nolimits_{c = 1}^{C} y_{o,c} \cdot log(1-p_{o,c} / 2).  
\end{equation}

By minimizing Eq.(\ref{equ:UCE_Loss}), we can drive the predicted probability $p_{o,c}$ to be closer to $0$ (as seen in Fig.~\ref{fig:ce_loss}(b)), thereby diminishing the prediction ability of the model on the target client's data to unlearn it. Note that before unlearning, the model $\omega^0$ often performs well on clients, where $p_{o,c}$ is close to $1$ and the model update step for the global model is already quite small. Hence, the constant ``2" in Eq.(\ref{equ:UCE_Loss}) is set to ensure that the gradient norm of the target client does not exceed those of the remaining clients. This can prevent the unlearning process from being unstable or even directly damaging the model utility. We verify this in Appendix.B.2.

Hence, when client $u$ requests for unlearning, it no longer applies GA on the CE loss. Instead, it switches to utilize UCE loss and performs gradient descent to train the model. Since UCE loss has the lower bound $0$, it can achieve the goal of unlearning client $u$'s data without bringing issues of gradient explosion and convergence difficulties. Denote $\tilde{L}_u$ as the local objective of the target client $u$ by using UCE loss, then the unlearning formulation (\ref{Problem:unlearning_objective}) is transferred to:
\begin{equation}
	\label{Problem:UCE_unlearning_objective}
	\min\limits_\omega~ \tilde{L}_u(\omega).
\end{equation}

\subsection{Orthogonal Steepest Descent Direction} \label{sec:OSD_direction}
In FedOSD, we solve Problem (\ref{Problem:UCE_unlearning_objective}) to unlearn the target client $u$ by iterating $\omega^{t+1} = \omega^t + \eta^t d^t$, where $d^t$ is an orthogonal steepest descent direction at $t^{th}$ round. In this section, we discuss how to obtain such an update direction and analyze how it can accelerate unlearning while mitigating the negative impact on the model utility. We start by introducing the gradient conflict, which is a direct cause of model performance degradation on FL clients \cite{FedFV}.

\textbf{Definition 1 (Gradient Conflict):}
The gradients of client $i$ and $j$ are in conflict with each other iff $g_i\cdot g_j < 0$.

In each communication round $t$, denote $g_i^t, i\neq u$ as the local gradient of remaining clients, and $g_u^t$ as the gradient of the target client $u$ for unlearning. If we directly adopt $-g_u^t$ as the direction to update the model to unlearn client $u$, i.e., $\omega^{t+1} = \omega^t - \eta^t g_u^t$, the model performance on the remaining clients would easily suffer reduction because $g_u^t$ would conflict with some $g_i^t$. The experimental results of Table~\ref{tab:exp_conflict} corroborate the presence and the impact of such gradient conflicts in FU.

Hence, mitigating gradient conflicts can help alleviate decreases in the model utility. One ideal solution would be identifying a common descent direction $d^t$ that satisfies $d^t$$\cdot$$g_u^t$$<$$0$ and $d^t$$\cdot$$g_i^t$$<$$0$. However, such a strategy could lead to the model becoming prematurely trapped in a local Pareto optimum, which remains far from the optimum of Problem (\ref{Problem:UCE_unlearning_objective}). We verify it in the ablation experiments (Section \ref{sec:ablation}).

To this end, we mitigate the gradient conflict by computing a model update direction $d^t$ orthogonal to the gradient of the remaining clients, i.e., $d^t\cdot g_i^t = 0, \forall i\neq u$. Although $d^t$ is not a common descent direction, it helps slow down the performance reduction of the model on the remaining clients. However, in FL, the number of remaining clients (i.e., $m-1$) is significantly smaller than $D$ (the dimension of model parameters), implying rank$(\forall g_i^t, i\neq u) \leq m-1 << D$. Consequently, there are numerous orthogonal vectors $d$ that satisfy $d \cdot g_i^t=0$. Therefore, if the obtained direction differs significantly from $-g_u^t$, it would impede the unlearning process and potentially exacerbate the degradation of model utility. We verify it in the ablation study in Section \ref{sec:ablation}.

Denote $G\in \mathbb{R}^{(m-1) \times D}$ as a matrix where each row represents a gradient of a remaining client, the key idea is to find a $d^t$ that satisfies $G d^t = \vec{0}$ while being closest to $-g_u^t$ to accelerate unlearning, i.e., $d^t = \mathop{\arg\min}\nolimits_{d^t}\mathop{cos}(g_u^t, d^t)$. To maintain the direction's norm, we fix $\|d^t\|$$=$$\|g_u^t\|$, then the problem is equivalent to:
\begin{equation}
	\label{Problem:orthogonal_steepest_descent_direction}
	\begin{array}{l}
		\mathop {\min }\limits_{d^t\in\mathbb{R}^D}~\frac{g_u^t\cdot d^t}{\|g_u^t\|^2}, \\
		\begin{aligned}
			s.t.~~ G d^t &=\vec{0},\\
			\|d^t\| &= \|g_u^t\|,
		\end{aligned}
	\end{array}
\end{equation}
which is a linear optimization problem and the solution is:
\begin{equation}
	\label{equ:osd_direction}
	d^t = \frac{1}{2\|g_u^t\|^2\mu}\left(G^TU\Sigma^+V^TGg_u^t - g_u^t\right),
\end{equation}
where $\mu$ is a related scalar that can make $\|d^t\|$$=$$\|g_u^t\|$, i.e., $\mu = \|G^TU\Sigma^+V^TGg_u^t - g_u^t\| / (2\|g_u^t\|^4)$. The matrices $V,\Sigma,U^T$ are the singular value decomposition of $GG^T$$\in$$\mathbb{R}^{(m-1)\times (m-1)}$, i.e., $GG^T = V\Sigma U^T$, which are not time-consuming to obtain. $\Sigma^+$ is the Moore-Penrose pseudoinverse of $\Sigma$, i.e., $\Sigma^+$ = diag$(\frac{1}{s_1}, \frac{1}{s_2},\cdots, \frac{1}{s_r}, 0, \cdots, 0)$, where $s_1, s_2, \cdots, s_r$ are the non-zero singular values of $GG^T$. The detailed proof of Eq.(\ref{equ:osd_direction}) is presented in Appendix.A.1, where we also report the actual computation time of FedOSD. The obtained $d^t$ is closest to $-g_u^t$ and satisfies $Gd^t=0$, so that it can accelerate unlearning and mitigate the model utility reduction.

\subsection{Gradient Projection in Post-training} \label{sec:post-training}
After unlearning, the target client $u$ leaves the FL system, and the remaining clients undertake a few rounds of FL training to recover the model utility. This phase is referred to as the ``post-training" stage \cite{FUPGA,MoDe}. However, we observe that not only is the model performance across remaining clients recovered, but unexpectedly, the performance on the forgotten data of the target client $u$ also improves. It looks like the model remembers what has been forgotten.

\begin{figure}[t!]
	\centering
	\includegraphics[width=0.8\columnwidth]{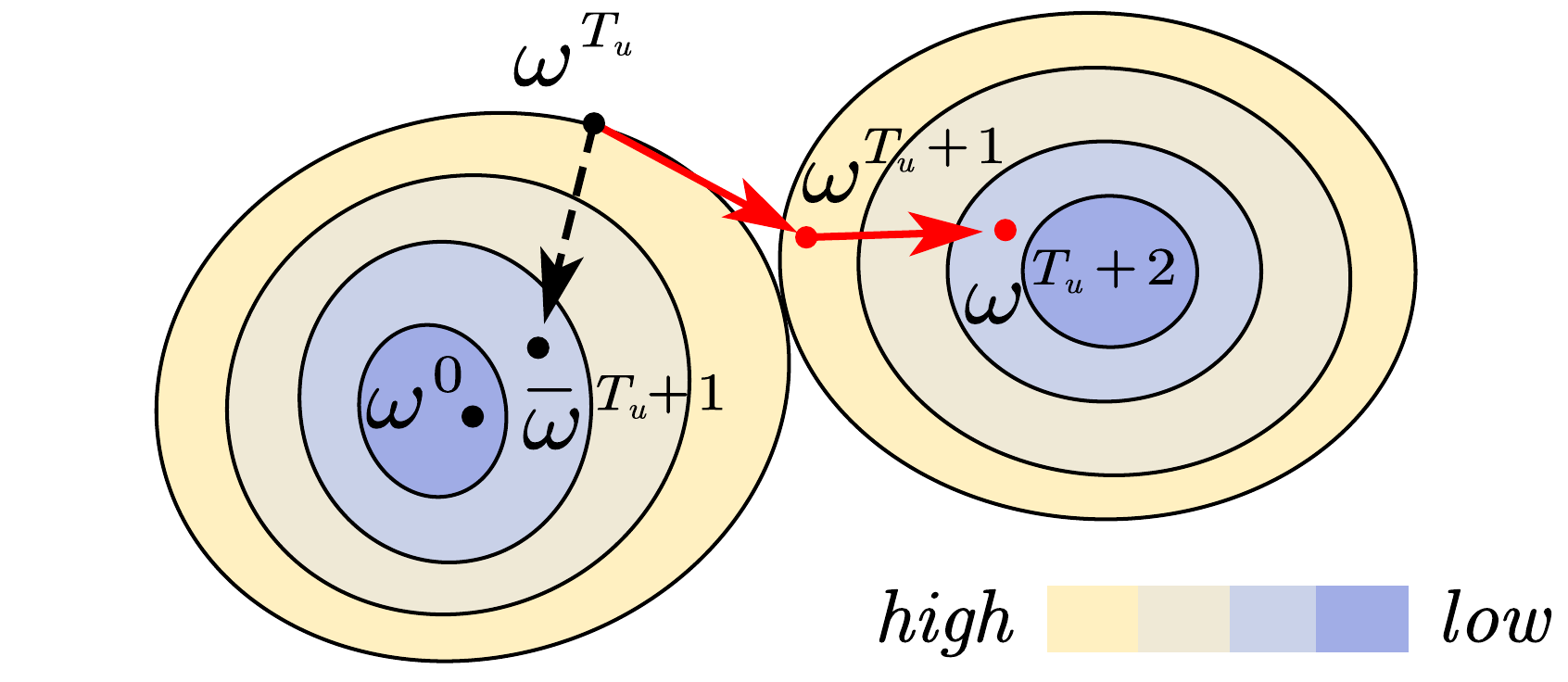}
	\caption{A demo depicting the model reverting issue in post-training. The contour map denotes the local loss of the model on a remaining client. $\omega^0$ is the original model before unlearning. $\omega^{T_u}$ is the model after unlearning. The dashed arrow depicts the path of the model update in post-training, where $\omega^{T_u}$ moves to $\bar \omega^{T_u+1}$ and is closer to $\omega^0$. The red arrows indicate a better path obtained by FedOSD.}
	\label{fig:reverting}
\end{figure}

One possible case is that the data from the target client $u$ share a similar distribution with the remaining clients' data. Hence, with the model utility being recovered, the model can generalize to client $u$'s data, thereby enhancing the model performance on client $u$. In general, this issue does not require intervention, because it can even happen on a retrained model without the participation of the target client $u$.

However, we observe that there is another case called model reverting that requires intervention. As seen in Fig.~\ref{fig:reverting}, with the model utility being reduced, the local loss of the remaining clients increased after unlearning. Besides, many previous FU algorithms do not significantly deviate the model from the original model $\omega^0$ during unlearning. Subsequently, when starting post-training, the local gradient $g_i^t$ does not conflict with $g_a^t$ (i.e., $g_i^t \cdot g_a^t > 0$), where $g_a^t$ is defined by $g_a^t=\nabla_{\omega^t} \frac{1}{2}\|\omega^t - \omega^0\|^2$. Therefore, the model is driven back to the old local optimal region where $\omega^0$ also resides, so that the model directly recovers what has been forgotten. The experimental results of Table~\ref{tab:exp1_unlearning_utility} and Fig.~\ref{fig:exp2} substantiate this observation, showing a decreased distance between the model and $\omega^0$ during post-training.

To address this issue, when $g_i^t \cdot g_a^t > 0$, we project the local gradient $g_i^t$ to the normal plane of $g_a^t$:
\begin{equation}
	\label{equ:project_ga}
	{g'}_i^t = g_i^t - \frac{g_i^t \cdot g_a^t}{\|g_a^t\|^2} \cdot g_a^t.
\end{equation}

Subsequently, each remaining client uploads $g_i'^t$ instead of $g_i^t$ to the server for the aggregation, i.e., $\bar g'^t = \frac{1}{|S|}\sum\nolimits_{i} g_i'^t$. And the global model is updated by $\omega^{t+1} = \omega^t - \eta \bar g'^t$. Given that ${g'}_i^t \cdot g_a^t = 0, \forall i$, $\bar g'^t$ satisfies $\bar g'^t \cdot g_a^t = 0$, ensuring that the updated model would not revert towards $\omega^0$. Thus, it addresses the reverting issue in the post-training stage. It's worth noting that the gradient projection method still works when the loss surface is complex. This is because it can always identify a direction that prevents the model from reverting to the original model, while guiding it towards other local optima.

\section{Experiments}
\begin{table*}[t!]
	\centering
	\resizebox{1\linewidth}{!}{
		\begin{tabular}{l|l|c|l|c|l|c|l|c|l|c|l|c}
			\toprule
			\multicolumn{1}{l|}{}       & \multicolumn{6}{c|}{FMNIST}
			& \multicolumn{6}{c}{CIFAR-10} \\ 
			\midrule
			\multicolumn{1}{l|}{}       & \multicolumn{2}{c|}{Pat-20}                              & \multicolumn{2}{c|}{Pat-50}                              & \multicolumn{2}{c|}{IID}                                & \multicolumn{2}{c|}{Pat-20}                              & \multicolumn{2}{c|}{Pat-50}                              & \multicolumn{2}{c}{IID}                                \\
			\midrule
			\multicolumn{1}{l|}{Algorithm} & \multicolumn{1}{c|}{ASR} & \multicolumn{1}{c|}{R-Acc} & \multicolumn{1}{c|}{ASR} & \multicolumn{1}{c|}{R-Acc} & \multicolumn{1}{c|}{ASR} & \multicolumn{1}{c|}{R-Acc} & \multicolumn{1}{c|}{ASR} & \multicolumn{1}{c|}{R-Acc} & \multicolumn{1}{c|}{ASR} & \multicolumn{1}{c|}{R-Acc} & \multicolumn{1}{c|}{ASR} & \multicolumn{1}{c}{R-Acc} \\ 
			\midrule
			$\omega^0$           & .991    & .852(.113)    & .957    & .869(.013)    & .893   & .898(.010)   & .897    & .589(.115)    & .754    & .658(.016)    & .243   & .731(.013)   \\
			Retraining   & .004    & .760(.228)    & .002    & .817(.025)    & .002   & .840(.015)   & .047    & .507(.106)    & .009    & .583(.149)    & .022   & .511(.013)   \\
			FedEraser    & .005    & .763(.171)    & .011    & .810(.102)    & .002   & .872(.009)   & .098    & .454(.158)    & .026    & .571(.128)    & .016   & .683(.012)   \\
			\midrule
			FedRecovery$^1$  & .637    & .761(.279)    & .693    & .823(.092)    & .498   & .871(.012)   & .156    & .454(.337)    & .102    & .476(.346)    & .015   & .692(.017)   \\
			MoDe$^1$         & .003    & .667(.246)    & .005    & .777(.046)    & .002   & .792(.012)   & .145    & .256(.162)    & .066    & .199(.119)    & .025   & .481(.018)   \\
			EWCSGA$^1$       & .000        & .255(.259)    & .000        & .233(.261)    & .101       & .126(.009)   & .000        & .199(.372)    & .000        & .381(.426)    & .018   & .259(.010)   \\
			FUPGA$^1$        & .000        & .227(.254)    & .000        & .178(.200)    & .101       & .105(.008)   & .000        & .202(.373)    & .000        & .388(.433)    & .019   & .271(.013)   \\
			FedOSD$^1$       & \textbf{.000}        & \textbf{.757(.187)}    & \textbf{.000}        & \textbf{.806(.042)}    & \textbf{.000}       & \textbf{.884(.011)}   & \textbf{.000}        & \textbf{.549(.185)}    & \textbf{.000}        & \textbf{.602(.175)}    & \textbf{.000}       & \textbf{.696(.016)}   \\
			\midrule
			FedRecovery$^2$ & .960$^r$     & .857(.112)    & .873$^r$    & .876(.013)    & .806$^r$   & .898(.011)   & .785$^r$    & .607(.119)    & .598$^r$    & .643(.138)    & .155$^r$   & .737(.016)   \\
			MoDe$^2$        & .007    & .744(.252)    & .003    & .816(.028)    & .002   & .843(.014)   & .060     & .519(.117)    & .035    & .582(.173)    & .016   & .703(.016)   \\
			EWCSGA$^2$      & .935$^r$    & .836(.173)    & .400$^r$      & .869(.013)    & .378$^r$   & .896(.012)   & .581$^r$    & .591(.194)    & .592$^r$    & .652(.118)    & .140$^r$   & .736(.016)   \\
			FUPGA$^2$       & .857$^r$    & .837(.185)    & .745$^r$    & .875(.013)    & .199$^r$   & .894(.009)   & .662	$^r$    & .599(.157)    & .602$^r$    & .658(.091)    & .144$^r$   & .737(.014)   \\
			FedOSD$^2$      & .023    & .851(.105)    & .021    & .874(.014)     & .004   & .897(.011)   & .027    & .606(.101)    & .016    & .659(.017)    & .030   & .734(.015)   \\
			\bottomrule
		\end{tabular}
	}
	\caption{The ASR, the mean R-Acc (and the std.) of the model. The row of $\omega^0$ denotes the initial state before unlearning. The `1' marked following the algorithm name represents the results after unlearning, while `2' denotes the results after post-training. The signal `$r$' in the columns of ASR ignifies an increase of the ASR value because of the model reverting during post-training.}
	\label{tab:exp1_unlearning_utility}
\end{table*}

\begin{table}[t!]
	\centering
	\resizebox{1\linewidth}{!}{
		\begin{tabular}{l|l|c|l|c|l|c}
			\toprule
			& \multicolumn{2}{c|}{Pat-20} & \multicolumn{2}{c|}{Pat-50} & \multicolumn{2}{c}{IID} \\
			\midrule
			& ASR        & R-Acc         & ASR        & R-Acc         & ASR      & R-Acc        \\
			\midrule
			SFU$^1$    & .000          & .345(.27)     & .198      & .218(.02)     & .103    & .169(.01)    \\
			\midrule
			SFU$^2$ & .547$^r$      & .792(.18)     & .563$^r$      & .846(.02)     & .386$^r$    & .893(.01)   \\
			\bottomrule
		\end{tabular}
	}
	\caption{The performance of SFU on FMNIST in Pat-20, Pat-50, and IID scenarios. All settings are the same as Table ~\ref{tab:exp1_unlearning_utility}. SFU$^1$ represents the results after unlearning, while SFU$^2$ denotes the results after post-training. The signal `$r$' in the columns of ASR signifies an increase in the ASR value because of the model reverting during post-training.}
	\label{tab:sfu}
\end{table}

\subsection{Experimental Setup}
We adopt the model test accuracy on the retained clients (denoted as R-Acc) to evaluate the model utility. To assess the effectiveness of unlearning, we follow \cite{FUPGA,SFU,MoDe} to implant backdoor triggers into the model by poisoning the target client's training data and flipping the labels (more details can be seen in Appendix.B.1). As a result, the global model becomes vulnerable to the backdoor trigger. The accuracy of the model on these data measures the attack success rate (denoted as ASR), and the low ASR indicates the effective unlearning performance of the algorithm.

\textbf{Baselines and Hyper-parameters.}
We first consider the retraining from scratch (denoted as Retraining) and FedEraser \cite{FedEraser}, which is also a kind of retraining but leverages the norms of the local updates stored in the preceding FL training to accelerate retraining. We then encompass well-known FU algorithms including FedRecovery \cite{FedRecovery}, MoDe \cite{MoDe}, and the gradient-ascent-based FU methods: EWCSGA \cite{EWCSGA} and FUPGA \cite{FUPGA}. We follow the settings of \cite{FUPGA,FedRecovery} that all clients utilize Stochastic Gradient Descent (SGD) on local datasets with local epoch $E=1$. We set the batch size as 200 and the learning rate $\eta \in \{0.005, 0.025, 0.001, 0.0005\}$ decay of 0.999 per round, where the best performance of each method is chosen in comparison. Prior to unlearning, we run FedAvg \cite{FedAvg} for 2000 communication rounds to generate the original model $\omega^0$ for unlearning. The max unlearning round is 100, while the max total communication round (including unlearning and post-training) is 200.

\textbf{Datasets and Models.}
We follow \cite{MoDe} to evaluate the algorithm performance on the public datasets MNIST \cite{MNIST}, FMNIST \cite{fashion}, and CIFAR-10/100 \cite{cifar}, where the training/testing data have already been split. To evaluate the effectiveness of unlearning across varying heterogeneous local data distributions, we consider four scenarios to assign data for clients: (1) Pat-20: We follow \cite{FedAvg} to build a pathological non-IID scenario where each client owns the data of 20\% classes. For example, in a dataset like MNIST with 10 classes, each client has two classes of the data. (2) Pat-50: It constructs a scenario where each client has 50\% classes. (3) Pat-10: It's an extreme data-island scenario where each client has 10\% of distinct classes. (4) IID: The data are randomly and equally separated among all clients. We utilize LeNet-5 \cite{MNIST} for MNIST, Multilayer perception (MLP) \cite{MLP} for FMNIST, CNN \cite{FUPGA} with two convolution layers for CIFAR-10, and NFResNet-18 \cite{NFResnet} for CIFAR-100.

\begin{table}[t]
	\resizebox{1\linewidth}{!}{
		\begin{tabular}{l|c|c|c|c|c|c}
			\toprule
			\multicolumn{1}{l|}{}   & \multicolumn{3}{c|}{MNIST}   & \multicolumn{3}{c}{CIFAR-100} \\
			\midrule
			\multicolumn{1}{l|}{Algorithm} & \multicolumn{1}{c|}{ASR} & \multicolumn{1}{c|}{R-Acc} & \multicolumn{1}{c|}{$NC$} & \multicolumn{1}{c|}{ASR} & \multicolumn{1}{c|}{R-Acc} & \multicolumn{1}{c}{$NC$} \\
			\midrule
			$\omega^0$                              & .997                    & .963                      & -                       & .584                    & .394                      & -                       \\
			\midrule
			FedRecovery                     & .038                    & .716                      & 3.00                       & .000                        & .214                      & 1.00                       \\
			MoDe                            & .039                    & .723                      & 3.47                    & .027                    & .160                       & 3.13                    \\
			EWCSGA                          & .000                        & .527                      & 7.45                    & .000                        & .093                      & 7.95                    \\
			FUPGA                           & .000                        & .535                      & 7.38                    & .000                        & .090                       & 7.92                    \\
			FedOSD                          & \textbf{.000}                        & \textbf{.924}                      & \textbf{0.00}                       & \textbf{.000}                        & \textbf{.369}                      & \textbf{0.00}                       \\
			\bottomrule
		\end{tabular}
	}
	\caption{ASR, R-Acc, and $NC$, the mean number of retained clients per round whose gradients conflict with the model update direction in Pat-20 on MNIST and CIFAR-100.}
	\label{tab:exp_conflict}
\end{table}

\begin{figure*}[t!]
	\centering
	\includegraphics[width=1.0\linewidth]{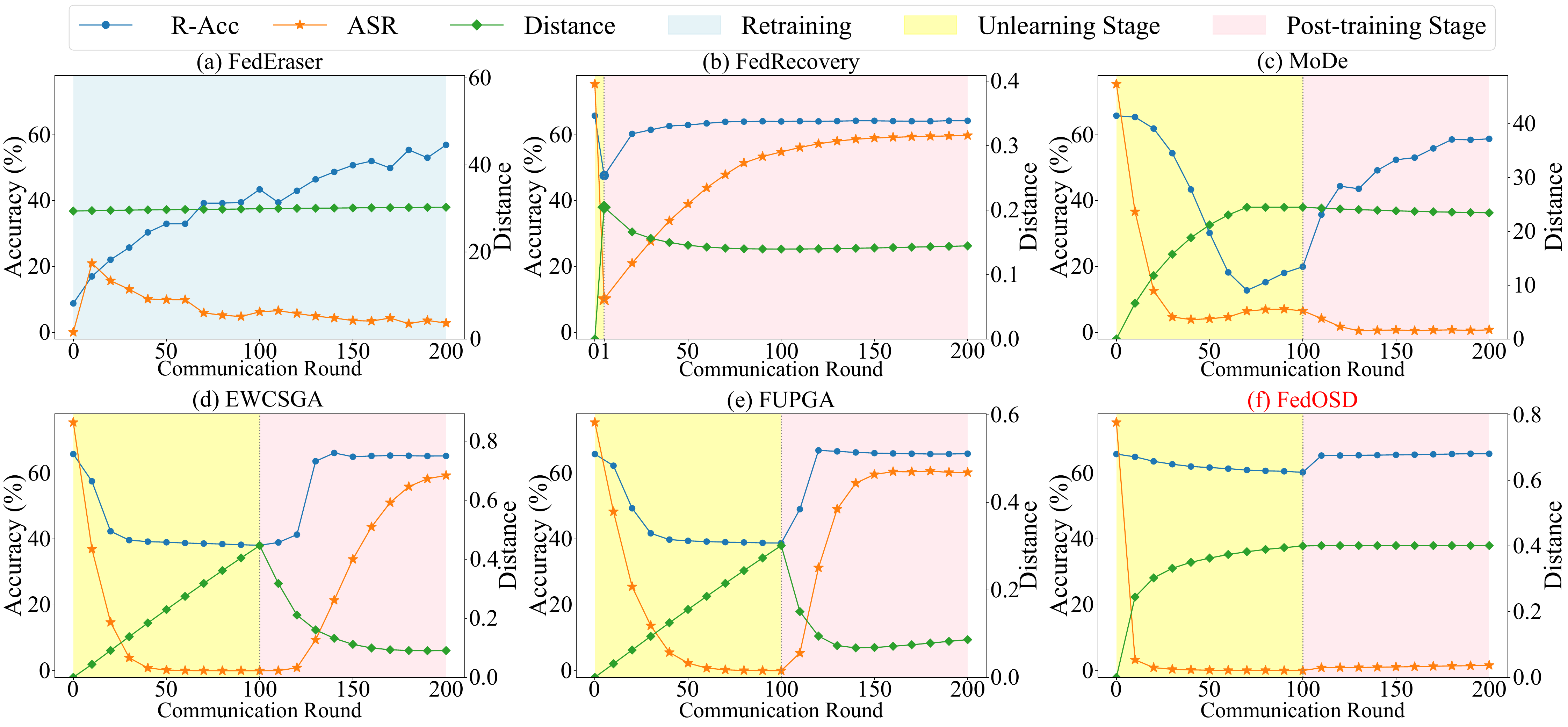}
	\caption{The ASR, the mean R-Acc, and the distance away from $\omega^0$ during unlearning and post-training stages in the Pat-50 scenario on CIFAR-10.}
	\label{fig:exp2}
\end{figure*}

\begin{table*}[t!]
	\centering
	\resizebox{0.92\linewidth}{!}{
		\begin{tabular}{l|l|c|c|c|l|c|c|c|l|c|c|c}
			\toprule
			\multicolumn{1}{l|}{}          & \multicolumn{4}{c|}{$m=10$}                                                                                        & \multicolumn{4}{c|}{$m=20$}                                                                                        & \multicolumn{4}{c}{$m=50$}                                                                                        \\ 
			\midrule
			\multicolumn{1}{l|}{Algorithm} & \multicolumn{1}{l|}{ASR} & \multicolumn{1}{c|}{R-Acc} & \multicolumn{1}{c|}{Worst} & \multicolumn{1}{c|}{Best} & \multicolumn{1}{l|}{ASR} & \multicolumn{1}{c|}{R-Acc} & \multicolumn{1}{c|}{Worst} & \multicolumn{1}{c|}{Best} & \multicolumn{1}{l|}{ASR} & \multicolumn{1}{c|}{R-Acc} & \multicolumn{1}{c|}{Worst} & \multicolumn{1}{c}{Best} \\ 
			\midrule
			$\omega^0$                              & .754                    & .658(.016)                  & .629  & .683 & .226                    & .678(.019)                  & .644                      & .712                     & .085                    & .712(.069)                  & .570                       & .830                      \\
			Retraining                      & .009                    & .583(.149)                  & .435  & .770  & .006                    & .499(.227)                  & .260                       & .768                     & .069                    & .408(.272)                  & .090                       & .730                      \\
			FedEraser                       & .026                    & .571(.128)                  & .399                      & .693                     & .045                    & .441(.253)                  & .158                      & .712                     & .018                    & .437(.057)                  & .325                      & .555                     \\
			\midrule
			FedRecovery$^1$                     & .102                    & .476(.346)                  & .082                      & .794                     & .012                    & .548(.115)                  & .400                        & .696                     & .000                        & .344(.348)                  & .000      & .740  \\
			MoDe$^1$                            & .066                    & .199(.119)                  & .062                      & .321                     & .026                    & .121(.063)                  & .054                      & .214                     & .116                    & .114(.056)                  & .035                      & .205                     \\
			EWCSGA$^1$                          & .000                        & .381(.426)                  & .000                          & .880                      & .000                        & .425(.439)                  & .000                          & .904                     & .000                        & .476(.329)                  & .085  & .850                      \\
			FUPGA$^1$                           & .000                        & .388(.433)                  & .000                          & \textbf{.889}                     & .000                        & .421(.443)                  & .000                          & \textbf{.914}                     & .000                        & .435(.368)                  & .045                      & \textbf{.855}                     \\
			FedOSD$^1$                          & \textbf{.000}                        & \textbf{.602(.175)}                  & \textbf{.433}                      & .803                     & \textbf{.000}                        & \textbf{.658(.103)}                  & \textbf{.526}                      & .798                     & \textbf{.000}                        & \textbf{.707(.067)}                  & \textbf{.575}                      & .830                      \\
			\midrule
			FedRecovery$^2$                    & .598$^r$                    & .643(.138)                  & .477  & .785 & .074$^r$                    & .667(.075)                  & .552                      & .766                     & .009$^r$                    & .644(.064)                  & .515  & .770  \\
			MoDe$^2$                           & .035                    & .582(.173)                  & .361                      & .747                     & .026                    & .503(.141)                  & .328                      & .666                     & .075                    & .368(.286)                  & .055                      & .710                      \\
			EWCSGA$^2$                         & .592$^r$                    & .652(.118)                  & .497                      & .781                     & .223$^r$                    & .680(.042)                  & .610                       & .752                     & .052$^r$                    & .711(.028)                  & .635                      & .780                      \\
			FUPGA$^2$                          & .602$^r$                    & .658(.091)                  & .538                      & .766                     & .220$^r$                     & .679(.041)                  & .614                      & .748                     & .046$^r$                    & .708(.029)                  & .640                       & .775                     \\
			FedOSD$^2$                         & .016                    & .659(.017)                  & .627  & .689 & .005                    & .678(.029)                  & .618                      & .730                      & .002                    & .710(.050)                  & .595                      & .820                     \\
			\bottomrule
		\end{tabular}
	}
	\caption{The ASR, the mean R-Acc (and the std.), as well as the worst and best R-Acc across the remaining clients on CIFAR-10 with Pat-50 under $m=10$, $m=20$, and $m=50$. The row of $\omega^0$ denotes the initial state before unlearning. The `1' marked following the algorithm name represents the results after unlearning, while `2' denotes the results after post-training. The signal `$r$' in the columns of ASR signifies an increase in the ASR value because of the model reverting during post-training.}
	\label{tab:different_N}
\end{table*}

\subsection{Evaluation of Unlearning and Model Utility}
We first evaluate the ASR and R-Acc of the model at the end of both the unlearning stage and post-training stage on FMNIST and CIFAR-10. One of the ten clients is randomly selected as the target client requesting for unlearning.

Table.~\ref{tab:exp1_unlearning_utility} lists the comparison results. It can be seen that in the unlearning stage, the gradient-ascent-based FU algorithms such as EWCSGA and FUPGA achieve more complete unlearning in non-IID scenarios, evidenced by their ASR reaching 0, but they experience a more pronounced reduction in R-Acc. What's worse, on FMNIST, the models of EWCSGA and FUPGA after the unlearning stage are nearly equivalent to a randomly initialized model. This is because their gradient constraint mechanisms, which aim to handle the gradient explosion issue, rely on fixed hyper-parameters. Since the optimal hyper-parameters cannot be determined in advance, these methods inevitably become ineffective.
Benefiting from the UCE loss and the orthogonal steepest descent update direction, the proposed FedOSD does not bring extra hyper-parameters and can successfully unlearn the target client data while suffering less utility reduction than others. Besides, since FedRecovery performs unlearning relies solely on the pre-stored historical FL training information, it cannot guarantee the unlearning effect in all scenarios. 

During the post-training stage, FedRecovery, EWCSGA, and FUPGA can recover the R-Acc to a level comparable to or exceeding that of the initial state. However, their models gravitate towards the initial $\omega^0$, leading to the models remembering what has been erased, and thus the ASR values rise significantly. In comparison, FedOSD can recover the model utility without suffering the model reverting issue. More experimental results on MNIST and CIFAR-100 are available in Appendix.B.2.

We also replicate SFU \cite{SFU} discussed in Section \ref{sec:Federated Unlearning} and test its performance on FMNIST (see Table \ref{tab:sfu}). For each remaining client, one batch of data samples is selected to compute the representation matrix. However, we find this process to be highly time-consuming due to the high dimensionality of the representation matrix, which complicates the computation of the SVD. The results depict that it cannot achieve the unlearning goal, suffering significant R-Acc reduction during the unlearning process, as well as the model reverting issue during post-training.

Furthermore, we present the experimental results for different client numbers: $m=10$, $m=20$, and $m=50$ in Table~\ref{tab:different_N}. These results verify the superior performance of FedOSD in terms of the unlearning effectiveness and the model utility in scenarios with more client participation.

To elucidate the negative impact of gradient conflicts on the model utility during unlearning, we report ASR, R-Acc, and the average number of retained clients experiencing gradient conflicts with the model update direction $d^t$ in Table.~\ref{tab:exp_conflict}. The results demonstrate that mitigating the conflict between $d^t$ and the remaining clients' gradients can significantly alleviate reductions in the model utility.

\begin{figure}[t!]
	\centering
	\includegraphics[width=1\columnwidth]{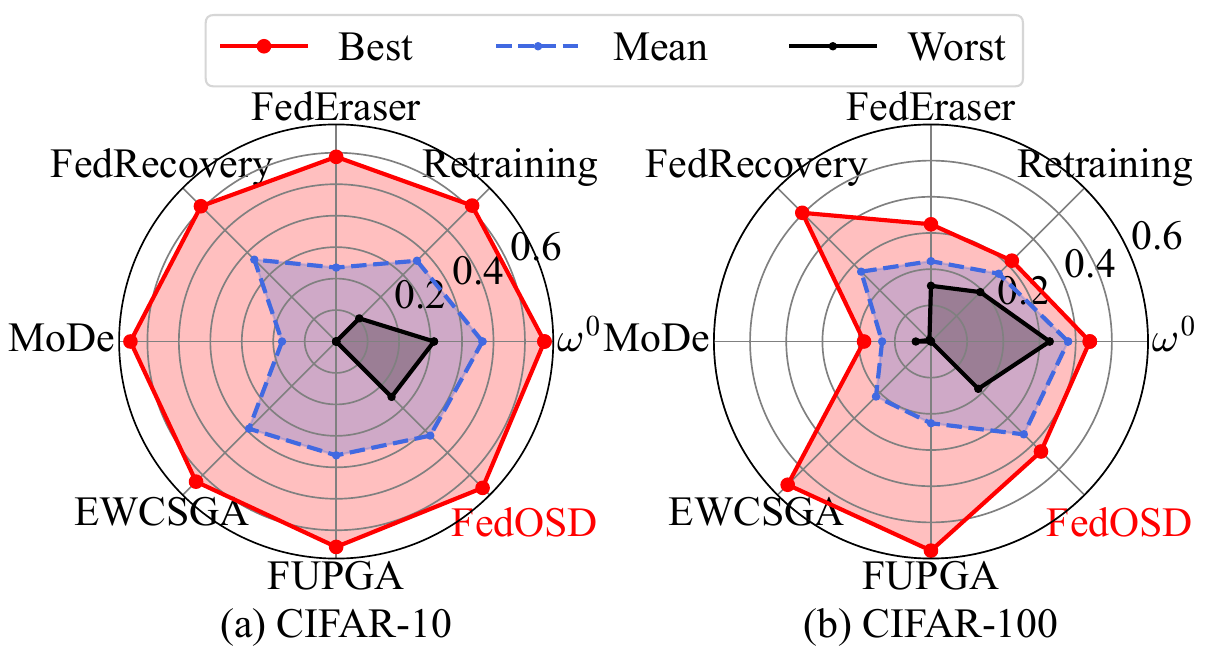}
	\caption{The best, the average, and the worst R-Acc across clients in Pat-10 on (a) CIFAR-10 and (b) CIFAR-100.}
	\label{fig:radar}
\end{figure}

Besides, Fig.~\ref{fig:radar} depicts the results in Pat-10 to evaluate the effect of unlearning on the model utility when some classes of data are completely removed. Compared with FedOSD, the model utility reduction on previous FU methods is considerably unfair, where the R-acc values are even approaching 0. In contrast, FedOSD more effectively maintains the model's performance on the remaining clients.

Moreover, we visualize the curves of ASR, R-Acc, and the distance between $\omega^t$ and $\omega^0$ during unlearning and post-training in Fig.~\ref{fig:exp2}. Notably, the unlearning stage of FedRecovery only comprises a single round, as it performs unlearning relying solely on the historical information of the previous FL training. The results demonstrate that FedOSD successfully achieves a zero ASR while maintaining the highest model utility during unlearning. The distance curve in the post-training stage verifies that the models of FedRecovery, EWCSGA, and FUPGA tend to revert towards $\omega^0$, evidenced by the decreasing distance, thereby leading to an increase in ASR, which suggests a recovery of previously unlearned information. In contrast, FedOSD prevents the model from moving back, thereby ensuring the recovery of model utility without suffering the model reverting issue during post-training.

\subsection{Ablation Experiments} \label{sec:ablation}
In Table.~\ref{tab:ablation}, we evaluate the performance of several variants of FedOSD (M1 to M5) to study the effect of each part.

\textbf{M1}: Do not use the UCE loss. Instead, the target client utilizes Gradient Ascent on the CE loss to unlearn. The results demonstrate that GA would destroy the model utility.

\textbf{M2}: Replace the orthogonal steepest descent direction $d^t$ to $-g_u^t$ for updating the unlearning model, which would conflict with retained clients' gradients. As a result, the model utility suffers more reduction than FedOSD.

\textbf{M3}: During unlearning, using Multiple Gradient Descent algorithm \cite{fliege2000steepest, FedLF} to obtain a common descent direction $d^t$ that satisfies $d^t$$\cdot$$g_i^t$$<$$0, \forall i\neq u$, which can both reduce the UCE loss of the target client and the CE loss of remaining clients in unlearning. The results depict that while this strategy does not compromise model utility, it fails to achieve the unlearning goal, verifying the analysis in Section \ref{sec:OSD_direction}.

\textbf{M4}: Randomly select a solution $d^t$ from the solutions to $G$$\cdot$$d^t$$=$$\vec{0}$ that also satisfies $d^t$$\cdot$$g_u^t$$<$$0$ to update the model for unlearning. Since the obtained $d^t$ would deviate a lot from $-g_u^t$, the result of ASR is higher than that of FedOSD. If we tune a larger learning rate to enhance the unlearning performance, it would further harm the model utility.

\textbf{M5}: Remove the gradient projection strategy in the post-training stage. It results in the model reverting issue, with a significant increase in ASR, verifying it's necessary to prevent the model from moving back to $\omega^0$ during post-training.

\begin{table}[t!]
	\centering
	\resizebox{1\linewidth}{!}{
		\begin{tabular}{l|l|c|l|c|l|c}
			\toprule
			\multicolumn{1}{l|}{}          & \multicolumn{2}{c|}{FMNIST}                              & \multicolumn{2}{c|}{CIFAR-10}                            & \multicolumn{2}{c}{CIFAR-100}                           \\ 
			\midrule
			\multicolumn{1}{l|}{Method} & \multicolumn{1}{l|}{ASR}  & \multicolumn{1}{l|}{R-Acc} & \multicolumn{1}{l|}{ASR}  & \multicolumn{1}{l|}{R-Acc} & \multicolumn{1}{l|}{ASR}  & \multicolumn{1}{l}{R-Acc} \\ 
			\midrule
			$\omega^0$                              & .957 & .869    & .754 & .658    & .433 & .437    \\
			\midrule
			FedOSD$^1$                          & \textbf{.000}     & .806     & \textbf{.000}     & .602    & \textbf{.000}     & .399    \\
			M1$^1$                              & .000                         & .163                        & .000                         & .224                        & .000                         & .014                        \\
			M2$^1$                              & .000                         & .317                        & .000                         & .391                        & .000                         & .267                        \\
			M3$^1$                              & .835                     & \textbf{.886}                        & .331                     & \textbf{.697}                        & .159                     & \textbf{.476}                        \\
			M4$^1$                              & .046                         & .693                        & .009                     & .566                        & .004                     & .360                        \\
			M5$^1$                              & .000     & .806    & .000                         & .602    & .000                         & .399                        \\
			\midrule
			FedOSD$^2$                          & \textbf{.021} & .874    & \textbf{.024} & .659    & \textbf{.056} & .458    \\
			M5$^2$                              & .244$^r$                     & .878                        & .340$^r$                      & .655                         & .138$^r$                     & .462 \\
			\bottomrule
		\end{tabular}
	}
	\caption{The ASR, the mean R-Acc of the model in the ablation studies. `1' marks the unlearning stage and `2' denotes post-training. `$r$' means suffering the model reverting issue.}
	\label{tab:ablation}
\end{table}

\section{Conclusion and Future Work}
In this work, we identify the convergence issue of Gradient Ascent and demonstrate the necessity of mitigating the gradient conflict in Federated Unlearning. Moreover, we highlight the issue of model reverting during post-training, which adversely affects the unlearning performance. To address these issues, we propose FedOSD, which modifies the Cross-Entropy loss to an unlearning version and achieves an orthogonal steepest descent model direction for unlearning. Extensive experiments verify that FedOSD outperforms SOTA FU methods in terms of the unlearning effect and mitigating the model utility reduction. A number of interesting topics warrant future exploration, including the design of the unlearning version of other loss functions such as MSE loss, and further enhancing fairness and privacy protection in FU.

\section*{Acknowledgments}
This work is supported in part by National Natural Science Foundation of China (72331009, 72171206, 62001412), in part by the Shenzhen Institute of Artificial Intelligence and Robotics for Society (AIRS), in part by Shenzhen Key Lab of Crowd Intelligence Empowered Low-Carbon Energy Network (No. ZDSYS20220606100601002), and in part by the Guangdong Provincial Key Laboratory of Future Networks of Intelligence (Grant No. 2022B1212010001).

\bibliography{aaai25}

\newpage

\appendix

\onecolumn

\section*{\huge Appendix}

\section{Theoretical Analysis and Proof}
In Section \ref{sec:osd_direction}, we begin by reviewing the proposed Unlearning Cross-Entropy loss, followed by an analysis of how FedOSD calculates the orthogonal steepest descent direction that is perpendicular to remaining clients' gradients while being closest to the gradient of the target unlearning client. In Section \ref{sec:Convergence Analysis}, we discuss the convergence of FedOSD during both the unlearning stage and the post-training stage, respectively. In Section \ref{sec:related_work}, we explore the related work and outline the distinctions between our approach and existing methodologies. Finally, in Section \ref{privacy}, we discuss the expectation for privacy protection within FedOSD.

\subsection{Theoretical Analysis of FedOSD} \label{sec:osd_direction}

\subsubsection{Unlearning Cross-Entropy Loss}
In Gradient Ascent (GA) based Federated Unlearning (FU), the target client aims to maximize the empirical risks over the local training data defined by Cross-Entropy Loss:
\begin{equation}
	\label{appendix_equ:CE_Loss}
	L_{CE} = -\sum\nolimits_{c = 1}^{C} y_{o,c} \cdot log(p_{o,c}),
\end{equation}
where $C$ represents the number of the data classes. $y_{o,c}$ is the binary indicator (0 or 1) if class label $c$ is the correct classification for observation $o$, i.e., the element of the one-hot encoding of sample $j$'s label. $p_{o,c}$ represents the predicted probability observation $o$ that is of class $c$, which is the $o^{th}$ element of the softmax result of the model output. Take a four-class classification task as an example. For a data sample with an actual label of 2, the corresponding one-hot encoded label vector $y=(0, 1, 0, 0)$, indicating that $y_{o,2}=1$. Suppose the softmax of the model output is $p=(0.05,0.8,0.05,0.1)$, then $p_{o,2}=0.8$ and the CE loss for this data sample is 0.223. The closer the $p_{o,c}$ is to 0, the larger the CE loss gets. Hence, the formulation of unlearning the target client $u$ can be defined by:
\begin{equation}
	\label{appendix_Problem:unlearning_objective}
	\max\limits_\omega~ L_u(\omega),
\end{equation}
where $L_u(\omega)$ represents the local objective of client $u$ in FL. However, since Eq.~(\ref{appendix_equ:CE_Loss}) has no upper bound, maximizing Problem (\ref{appendix_Problem:unlearning_objective}) by GA would directly lead to the gradient explosion, rendering the process non-convergent.

In FedOSD, rather than performing GA on the CE loss, the target client switches to utilize the proposed Unlearning Cross-Entropy (UCE) loss Eq.~(\ref{appendix_equ:UCE_Loss}) and performs gradient descent to drive $p_{o,c}$ to 0 to reduce the model performance on the target client's data to unlearn.
\begin{equation}
	\label{appendix_equ:UCE_Loss}
	L_{UCE} = -\sum\nolimits_{c = 1}^{C} y_{o,c} \cdot log(1-p_{o,c} / 2).  
\end{equation}

Therefore, in FedOSD, the formulation of unlearning the target client $u$ is transferred to
\begin{equation}
	\label{appendix_Problem:UCE_unlearning_objective}
	\min\limits_\omega~ \tilde{L}_u(\omega).
\end{equation}
where $\tilde{L}_u$ represents the local objective of the target client $u$ by using UCE loss.

\subsubsection{Orthogonal Steepest Descent Direction}
In FedOSD, we solve Problem (\ref{appendix_Problem:UCE_unlearning_objective}) to ease the target client $u$'s data by iterating $\omega^{t+1} = \omega^t + \eta^t d^t$ at each communication round $t$, where $\eta^t$ denotes the step size (learning rate) and $d^t$ is a direction for updating the model. To mitigate the gradient conflicts, $d^t$ is orthogonal to remaining clients' gradients $g_i^t$ while being closest to the inverse of the target client $u$'s gradient $g_u^t$:
\begin{equation}
	\label{appendix_Problem:cosine_direction}
	\begin{array}{l}
		\mathop {\max }\limits_{d^t\in\mathbb{R}^D}~{\rm{cos}}(-g_u^t, d^t), \\
		s.t.~~ G d^t =\vec{0},
	\end{array}
\end{equation}
where `cos' is the cosine similarity, and $D$ is the number of the model parameters. $G \in \mathbb{R}^{(m-1) \times D}$ denotes a matrix concatenated by remaining clients' gradients $g_i^t, \forall i \neq u$, i.e., each row of $G$ is a local gradient of remaining client $i$. To maintain the norm of the update direction, we fix $\|d^t\| = \|g_u^t\|$. Then, the above problem is written as:
\begin{equation}
	\label{appendix_Problem:orthogonal_steepest_descent_direction}
	\begin{array}{l}
		\mathop {\min }\limits_{d^t\in\mathbb{R}^D}~\frac{g_u^t\cdot d^t}{\|g_u^t\|^2}, \\
		\begin{aligned}
			s.t.~~ G d^t &=\vec{0},\\
			\|d^t\| &= \|g_u^t\|,
		\end{aligned}
	\end{array}
\end{equation}
From the Karush-Kuhn-Tucker (KKT) condition, there exist $\mu \in \mathbb{R}$, $\lambda \in \mathbb{R}^{m-1}$ and $d^*$ that satisfy:
\begin{equation}
	\label{KKT_1}
	\frac{g_u^t}{\|g_u^t\|^2} + 2\mu d^* + G^T \lambda = \vec{0},
\end{equation}
\begin{equation}
	\label{KKT_2}
	\mu \geq 0,
\end{equation}
\begin{equation}
	\label{KKT_3}
	\mu(\|d^*\|^2 - \|g_u^t\|^2) = 0,
\end{equation}
\begin{equation}
	\label{KKT_4}
	G d^t =\vec{0}.
\end{equation}

From (\ref{KKT_1}), we know that for some $\lambda \in \mathbb{R}^{m-1}$, we have:
\begin{equation}
	\label{temp_1}
	d^* = -\frac{1}{2\mu}\left(G^T\lambda + \frac{g_u^t}{\|g_u^t\|^2}\right).
\end{equation}

Substituting it to (\ref{KKT_4}), we have:
\begin{equation}
	\label{temp_2}
	-\frac{1}{2\mu}\left(GG^T\lambda + \frac{Gg_u^t}{\|g_u^t\|^2}\right) = \vec{0}.
\end{equation}

Solve it, we have $\mu = 0$ or:
\begin{equation}
	\label{temp_3}
	GG^T\lambda + \frac{Gg_u^t}{\|g_u^t\|^2} = \vec{0}.
\end{equation}

Given that $\mu = 0$ is not the desired solution, our attention shifts to Eq.~(\ref{temp_3}). This equation implies:
\begin{equation}
	\label{temp_4}
	\lambda = -\frac{1}{\|g_u^t\|^2} (GG^T)^{-1}Gg_u^t.
\end{equation}

Since rank($G$) $\leq (m-1)$, $GG^T \in \mathbb{R}^{(m-1) \times (m-1)}$ might not be invertible, we utilize Moore-Penrose pseudoinverse of $GG^T$:
\begin{equation}
	\label{temp_5}
	(GG^T)^+ = U\Sigma^+V^T,
\end{equation}
where $U,\Sigma,V$ are the singular value decomposition of $GG^T$, i.e., $GG^T = V\Sigma U^T$. Hence, 
\begin{equation}
	\label{temp_6}
	\lambda = -\frac{1}{\|g_u^t\|^2} U\Sigma^+ V^T Gg_u^t.
\end{equation}
Substituting (\ref{temp_6}) to (\ref{temp_1}), we obtain:
\begin{equation}
	\label{appendix_equ:osd_direction}
	d^* = \frac{1}{2\|g_u^t\|^2\mu}\left(G^TU\Sigma^+V^TGg_u^t - g_u^t\right).
\end{equation}
Finally, from $\|d^*\|^2 - \|g_u^t\|^2$, we can obtain:
\begin{equation}
	\label{cal_mu}
	\mu = \frac{\|G^TU\Sigma^+V^TGg_u^t - g_u^t\|}{2\|g_u^t\|^4}.
\end{equation}

Therefore, at each communication round $t$, we utilize $d^t = \frac{1}{2\|g_u^t\|^2\mu}\left(G^TU\Sigma^+V^TGg_u^t - g_u^t\right)$ as the model update direction. Since it is closest to $-g_u^t$ while being perpendicular to $g_i^t, \forall i \neq u$, this direction can accelerate unlearning while mitigating the gradient conflicts, and thus reduce the impact on model utility. Note that the process is not time-consuming as Eq.~(\ref{appendix_equ:osd_direction}) only contains few matrix multiplication, and Eq.~(\ref{temp_5}) can be quickly obtained since $GG^T\in \mathbb{R}^{(m-1) \times (m-1)}$. The actual computation time of FedOSD is presented in Table~\ref{runtime}.

\subsubsection{Gradient Projection in Post-training}
In the post-training stage \cite{FUPGA,FedKdu,MoDe}, the target client $u$ leaves the FL system, and the remaining clients undertake a few rounds of FL training to recover the model utility that was reduced in the previous unlearning stage. However, both theoretical and empirical analyses indicate that the FL global model easily suffers a model reverting issue in this stage, wherein the distance between the current model and the original model $\omega^0$ decreases. The model reverting issue directly prevents the model from recovering what has been unlearned before, thereby undermining the achievements of the unlearning process.

To this end, we maintain the unlearning achievement by introducing a gradient projection strategy. In the post-training stage, after each remaining client conducts their local training to obtain $\omega_i^t$, we compute $g_i^t = (\omega^t - \omega_i^t) / \eta^t$ and $g_a^t = \nabla_{\omega^t} \frac{1}{2}\|\omega^t - \omega^0\|^2$. If $g_i^t \cdot g_a^t > 0$, we project $g_i^t$ to the normal plane of $g_a^t$:
\begin{equation}
	\label{appendix_equ:project_ga}
	g_i'^t = g_i^t - \frac{g_i^t \cdot g_a^t}{\|g_a^t\|^2} \cdot g_a^t.
\end{equation}
Hence, the obtained $g_i'^t$ satisfies $g_i'^t \cdot g_a^t = 0$, ensuring that it does not contribute to moving the model closer to $\omega^0$. Afterwards, each remaining client $i$ uploads $g_i'^t$ to the server, and the model is updated by:
\begin{equation}
	\omega^{t+1} \leftarrow \omega^t - \eta^t \frac{1}{|S|}\sum\nolimits_{i} g_i'^t.
\end{equation}

\subsection{Convergence Analysis} \label{sec:Convergence Analysis}
In this section, we first prove the convergence and analyze the convergence rate of FedOSD in the unlearning stage, we then discuss the convergence in the post-training stage.

\subsubsection{Convergence in the Unlearning Stage}
We prove that FedOSD can converge in the unlearning stage as follows:

Assume that the local object $\tilde L_u(\omega)$ of the target client $u$ is differentiable and Lipschitz-smooth (L-smooth) with the Lipschitz constant $\mathcal{L}$.

Denote $g_u^t$ as the local gradient of the target client $u$ at round $t$. Denote $T_u$ as the given maximum communication round of unlearning. 

For $\tilde L(\omega^t)$, any $t_1, t_2$, since the angle between $-g_u^t$ and $d^t$ obtained by (\ref{appendix_equ:osd_direction}) is smaller than $90^\circ$, $d^t$ satisfies gradient descent for client $u$, i.e., for any $t_1 \neq t_2$:
\begin{equation}
	\label{eq:1}
	\tilde L_u({\omega ^{t_1}}) \geq \tilde L_u({\omega ^{t_2}}) + (-d^t) \cdot ({\omega ^{t_1}} - {\omega ^{t_2}}).
\end{equation}

Since $\tilde{L_u}$ is Lipschitz continuous,
\begin{equation}
	\label{eq:0}
	\tilde L_u({\omega ^{t_1}}) - \tilde L_u({\omega ^{t_2}}) \leq \mathcal{L}{\left. {\left\| {{\omega ^{t_1}} - {\omega ^{t_2}}} \right.} \right\|_2}.
\end{equation}

From Eq.~(\ref{eq:1}) we can get
\begin{equation}
	\label{eq:2}
	\tilde L_u({\omega ^{t - 1}}) - \tilde L_u({\omega ^*}) \ge  - {d^t} \cdot ({\omega ^{t - 1}} - {\omega ^*}).
\end{equation}

Besides, 
\begin{align}
	\left\| \omega^t - \omega^* \right\|_2^2 &= \left\| \omega^{t-1} + \eta^t d^t - \omega^* \right\|_2^2\notag\\
	&= \left\| \omega^{t-1} - \omega^* \right\|_2^2 + 2\eta^t  d^t \cdot (\omega^{t-1} - \omega^*) + ({\eta ^t})^2 \left\| d^t \right\|_2^2\notag\\
	&\le \left\| \omega^{t-1} - \omega^* \right\|_2^2 - 2\eta^t (L(\omega^{t-1}) - L(\omega^*)) + ({\eta ^t})^2 \left\| d^t \right\|_2^2.
\end{align}

The above formula implies
\begin{equation}
	\label{eq:5}
	\left\| {{\omega ^t} - } \right.\left. {{\omega ^*}} \right\|_2^2 \le \left\| {{\omega ^0}} \right. - \left. {{\omega ^*}} \right\|_2^2 - 2\sum\nolimits_{t = 1}^{T_u} {{\eta ^t}(\tilde L_u({\omega ^{t - 1}}) - \tilde L_u({\omega ^*}))}  + \sum\nolimits_{t = 1}^{T_u} {({\eta ^t})^2\left\| {{d^t}} \right\|_2^2}.
\end{equation}

Denote $R = \|\omega^0 - \omega^*\|_2$, obviously it has $R^2 \geq 0$. Hence,
\begin{equation}
	0 \leq R^2 - 2\sum\nolimits_{t = 1}^{T_u} {{\eta ^t}(\tilde L_u({\omega ^{t-1}}) - \tilde L_u({\omega ^*}))}  + \sum\nolimits_{i = 1}^n {({\eta ^t})^2}\left\| {{d^t}} \right\|_2^2.
\end{equation}

Introducing $\tilde L_u(\omega^{T_u}_{best}) = {\min _{t = 1...T_u}}\tilde L_u({\omega ^t})$ and substituting for $\tilde L_u (\omega^{t-1}) - \tilde L_u(\omega^*)$ makes the right side larger:
\begin{equation}
	\label{eq:6}
	0 \leq R^2 - 2\sum\nolimits_{t = 1}^{T_u} {{\eta ^t}(\tilde L_u({\omega^{T_u}_{best}}) - \tilde L_u({\omega ^*}))}  + \sum\nolimits_{i = 1}^{T_u} {({\eta ^t})^2}\left\| {{d^t}} \right\|_2^2.
\end{equation}

Afterwards, considering Eq.~(\ref{eq:0}) and Eq.~(\ref{eq:6}), we have
\begin{equation}
	\label{eq:7}
	\tilde L_u({\omega^{T_u}_{best}}) - \tilde L_u(\omega^*) \le \frac{R^2 + {\mathcal{L}^2}\sum\nolimits_{t = 1}^{T_u} {({\eta ^t})^2} }{{2\sum\nolimits_{t = 1}^{T_u} {{\eta ^t}} }},
\end{equation}
which implies
\begin{equation}
	\label{L_limit}
	\lim_{t\to\infty} ||\tilde L_u(\omega^t) - \tilde L_u(\omega^{*})||= 0.
\end{equation}

Therefore, in the unlearning stage, FedOSD can converge to the local optimum of Problem (\ref{Problem:UCE_unlearning_objective}).

\textbf{Convergence Rate.}

Consider the right hand side of Eq.~\ref{eq:7}, taking $\eta^{t} = \frac{R}{\mathcal{L}\sqrt{t}}$, $t=1,\cdots, T_u$. The basis bound is
\begin{equation}
	\frac{{{R^2} + {\mathcal{L}^2}\sum\nolimits_{t = 1}^{T_u} {{{({\eta ^t})}^2}} }}{{2\sum\nolimits_{t = 1}^{T_u} {{\eta ^t}} }} = \frac{{R\mathcal{L}}}{{\sqrt{T_u} }}.
\end{equation}

This means FedOD has convergence rate $O(\frac{1}{\sqrt{T_u}})$.

\subsubsection{Convergence in the Post-training Stage}
We prove that in the post-training stage, FedOSD can converge to the following FL objective:
\begin{equation}
	\label{FL_obj}
	\min\nolimits_\omega \sum\nolimits_{i = 1}^{|S|} \frac{1}{|S|} L_i(\omega).
\end{equation}

Denote $S$ as the set of remaining clients in the post-training stage, $L_i(\omega)$ as the local objective of remaining client $i \in S$, and the corresponding local gradient of client $i$ is $g_i$. Let $L^*$ and $L_i^*$ be the minimum values of $L$ and $L_i$.

\textbf{Assumption 1}. For all remaining client $i\in S$, the local objective $L_i(\omega)$ is differentiable and L-smooth with Lipschitz constant $\mathcal L_i$:
\begin{equation}
	\label{eq:2_1}
	L_i(\omega^{t_1}) \leq L_i(\omega^{t_2}) + (\omega^{t_1} - \omega^{t_2})^T g_i^{t_2} + \frac{\mathcal{L}_i}{2} \|\omega^{t_1} - \omega^{t_2}\|_2^2.
\end{equation} 

\textbf{Assumption 2}. $L_i, \forall i \in S$ are $\mu-$strongly convex:
\begin{equation}
	\label{eq:2_2}
	L_i(\omega^{t_1}) \geq L_i(\omega^{t_2}) + (\omega^{t_1} - \omega^{t_2})^T g_i^{t_2} + \frac{\mu}{2} \|\omega^{t_1} - \omega^{t_2}\|_2^2.
\end{equation}

\textbf{Assumption 3}. The expected squared norm of $g_i'^t$ is uniformly bounded, i.e., for $i=1,\cdots,|S|$:
\begin{equation}
	\label{eq:2_3}
	\mathbb{E}{\left\| {g_i'^t} \right\|^2} \le {K^2}.
\end{equation}

\textbf{Theorem 1}. The expected squared norm of the difference between $g_i'^t$ and $g_i^t$ is uniformly bounded by $\sqrt{2}\|g_i^t\|$ at each round $t$:
\begin{equation}
	\label{eq:2_4}
	\mathbb{E}{\left\| {g_i'^t} - {g_i^t}\right\|^2} \le 2{\left\| {g_i^t}\right\|^2}.
\end{equation}

Proof of Theorem 1:
Since in the post-training stage, $g_i'^t$ is the projection of $g_i^t$ on the normal plane of $g_a^t = \nabla_{\omega^t} \frac{1}{2}\|\omega^t - \omega^0\|^2$, after which it is rescaled to keep $\|g_i'^t\| = \|g_i^t\|$, hence, the maximum of $\| {g_i'^t} - {g_i^t}\|^2$ is $2 \|g_i^t\|^2$. So the Theorem 1 holds.

\textbf{Lemma 1}. Denote $\mathcal{L}$ as the largest Lipschitz constant among $\mathcal{L}_1, \cdots, \mathcal{L}_{|S|}$. By Assumption 1 and 2, if \({\eta ^t} \le \frac{1}{{4\mathcal{L}}}\), we have
\begin{equation}
	\label{eq:2_5}
	\mathbb{E}{\left\| {{\omega ^{t + 1}} - {\omega ^*}} \right\|^2} \le (1 - \mu {\eta ^t})\mathbb{E}{\left\| {{\omega ^t} - {\omega ^*}} \right\|^2} + {({\eta ^t})^2}\mathbb{E}{\left\| {\bar g'^t - \bar g^t} \right\|^2} + 6\mathcal{L}{({\eta ^t})^2}\tau  + 2\mathbb{E}\sum\limits_{i = 1}^{\left| S \right|} {\frac{1}{{\left| S \right|}}} {\left\| {{\omega ^t} - \omega_i^t} \right\|^2}
\end{equation}
where,
\(\tau  = {L^*} - \sum\nolimits_{i = 1}^{\left| S \right|} {\frac{1}{{\left| S \right|}}}L^*_i \ge 0\).

\textbf{Proof of Lemma 1}: Notice that \({\omega ^{t + 1}} = {\omega ^t} - {\eta ^t}{\bar g'^t}\), then

\begin{align}
	\label{eq:L1_0}
	\left\| \omega^{t + 1} - \omega^* \right\|^2 &= \left\| \omega^t - \eta^t \bar g'^t - \omega^* + \eta^t \bar g^t - \eta^t \bar g^t \right\|^2 \notag\\
	&= \left\| \omega^t - \omega^* - \eta^t \bar g^t \right\|^2 + 2\eta^t \langle \omega^t - \omega^* - \eta^t \bar g^t, \bar g^t - \bar g'^t \rangle + ({\eta ^t})^2 \left\| \bar g^t - \bar g'^t \right\|^2
\end{align}

Let \({A_1}\) = \({\left\| {{\omega ^t} - {\omega ^*} - {\eta ^t}{{\bar g}^t}} \right\|^2}\), and
\({A_2} = 2{\eta ^t}\langle {\omega ^t} - {\omega ^*} - {\eta ^t}{\bar g^t},{\bar g^t} - {\bar g'^t}\rangle \).

Note that \(\mathbb{E}(A_2) = 0\), just focus on bounding \(A_1\):
\begin{equation}
	\label{eq:L1_1}
	{\left\| {{\omega ^t} - {\omega ^*} - {\eta ^t}{{\bar g}^t}} \right\|^2} = {\left\| {{\omega ^t} - {\omega ^*}} \right\|^2} - 2{\eta ^t}\langle {\omega ^t} - {\omega ^*},{{\bar g}^t}\rangle  + {({\eta ^t})^2}{\left\| {{{\bar g}^t}} \right\|^2}.
\end{equation}

Let \(B_1\) = \(- 2{\eta ^t}\langle {\omega ^t} - {\omega ^*},{\bar g^t}\rangle\), \(B_2\) = \({({\eta ^t})^2}{\left\| {{{\bar g}^t}} \right\|^2}\).

By Assumption 2, it follows that
\begin{equation}
	\label{eq:L1_2}
	{\left\| {g_i'^t} \right\|^2} = {\left\| {g_i^t} \right\|^2} \le 2\mathcal{L}({L_i}(\omega _i^t) - {L^*})
\end{equation}
By Assumption 1 and Eq.~(\ref{eq:L1_2}), we have
\begin{equation}
	{B_2} = {({\eta ^t})^2}{\left\| {{\bar g^t}} \right\|^2} \le {({\eta ^t})^2}\sum\limits_{i = 1}^{\left| S \right|} {\frac{1}{{\left| S \right|}}} {\left\| {g_i'^t} \right\|^2} \le 2\mathcal{L}{({\eta ^t})^2}\sum\limits_{i = 1}^{\left| S \right|} {\frac{1}{{\left| S \right|}}} ({L_i}(\omega _i^t) - L^*_i).
\end{equation}

Note that
\begin{align}
	\label{eq:L1_3}
	B_1 &= - 2{\eta ^t} \langle \omega^t - \omega^*, \bar{g}^t \rangle \
	= - 2{\eta ^t} \sum\limits_{i = 1}^{\left| S \right|} \frac{1}{\left| S \right|} \langle \omega^t - \omega^*, g^t_i \rangle \notag\\
	&= - 2{\eta ^t} \sum\limits_{i = 1}^{\left| S \right|} \frac{1}{\left| S \right|} \langle \omega^t - \omega^t_i, g^t_i \rangle 
	- 2{\eta ^t} \sum\limits_{i = 1}^{\left| S \right|} \frac{1}{\left| S \right|} \langle \omega^t_i - \omega^*, g^t_i \rangle.
\end{align}

By Cauchy-Schwarz inequality and AM-GM inequality, we have
\begin{equation}
	\label{eq:L1_4}
	- 2\langle {\omega ^t} - \omega ^t_i,g^t_i\rangle  \leq \frac{1}{{{\eta ^t}}}{\left\| {\omega ^t - \omega ^t_i} \right\|^2} + {\eta ^t}{\left\| g^t_i \right\|^2}.
\end{equation}

By Assumption 2, we have
\begin{equation}
	\label{eq:L1_5}
	- \langle \omega ^t_i - {\omega ^*},g^t_i\rangle  \leq  - ({L_i}(\omega ^t_i) - {L_i}({\omega ^*})) - \frac{\mu }{2}{\left\| {\omega ^t_i - {\omega ^*}} \right\|^2}.
\end{equation}

By combining Eq.~(\ref{eq:L1_1}) Eq.~(\ref{eq:L1_3}) Eq.~(\ref{eq:L1_4}) Eq.~(\ref{eq:L1_5}), it follows that
\begin{align}
	\label{eq:L1_6}
	A_1 &= \left\| \omega^t - \omega^* - \eta^t \bar g^t \right\|^2 \notag \\
	&\le \left\| \omega^t - \omega^* \right\|^2 + 2\mathcal{L}({\eta ^t})^2 \sum\limits_{i = 1}^{\left| S \right|} \frac{1}{\left| S \right|} \left( L_i(\omega_i^t) - L_i^* \right) \notag \\
	&\quad + \eta^t \sum\limits_{i = 1}^{\left| S \right|} \frac{1}{\left| S \right|} \left( \frac{1}{\eta^t} \left\| \omega^t - \omega_i^t \right\|^2 + \eta^t \left\| g_i^t \right\|^2 \right) \notag \\
	&\quad - 2\eta^t \sum\limits_{i = 1}^{\left| S \right|} \frac{1}{\left| S \right|} \left( L_i(\omega_i^t) - L_i(\omega^*) + \frac{\mu}{2} \left\| \omega_i^t - \omega^* \right\|^2 \right)\notag\\
	&= \left( {1 - \mu {\eta ^t}} \right){\left\| {{\omega ^t} - {\omega ^*}} \right\|^2} + \sum\limits_{i = 1}^{\left| S \right|} {\frac{1}{{\left| S \right|}}} {\left\| {{\omega ^t} - \omega ^t_i} \right\|^2} \notag\\ 
	&\quad + 4\mathcal{L}{({\eta ^t})^2}\sum\limits_{i = 1}^{\left| S \right|} {\frac{1}{{\left| S \right|}}} ({L_i}(\omega _i^t) - L^*_i) - 2{\eta ^t}\sum\limits_{i = 1}^{\left| S \right|} {\frac{1}{{\left| S \right|}}} ({L_i}(\omega ^t_i) - {L_i}({\omega ^*}))
\end{align}

where we use Eq.~(\ref{eq:L1_2}) again.

Let \(C = 4\mathcal{L}{({\eta ^t})^2}\sum\limits_{i = 1}^{\left| S \right|} {\frac{1}{{\left| S \right|}}} ({L_i}(\omega _i^t) - L^*_i) - 2{\eta ^t}\sum\limits_{i = 1}^{\left| S \right|} {\frac{1}{{\left| S \right|}}} ({L_i}(\omega ^t_i) - {L_i}({\omega ^*))}\).

Next step, we aim to bound $C$. We define \({\gamma ^t} = 2{\eta ^t}(1 - 2\mathcal{L}{\eta ^t})\). Since \({\eta ^t} \le \frac{1}{4\mathcal{L}}\), \({\eta ^t} \le {\gamma ^t} \le 2{\eta ^t}\). Then we split $C$ into:
\begin{align}
	\label{eq:L1_7}
	C &= - 2\eta^t(1 - 2\mathcal{L}\eta^t) \sum\limits_{i = 1}^{\left| S \right|} \frac{1}{\left| S \right|} \left( L_i(\omega_i^t) - L_i^* \right) + 2\eta^t \sum\limits_{i = 1}^{\left| S \right|} \frac{1}{\left| S \right|} \left( L_i(\omega^*) - L_i^* \right) \notag\\
	&= - \gamma^t \sum\limits_{i = 1}^{\left| S \right|} \frac{1}{\left| S \right|} \left( L_i(\omega_i^t) - L^* \right) + \left( 2\eta^t - \gamma^t \right) \sum\limits_{i = 1}^{\left| S \right|} \frac{1}{\left| S \right|} \left( L^* - L_i^* \right) \notag\\
	&= - \gamma^t \sum\limits_{i = 1}^{\left| S \right|} \frac{1}{\left| S \right|} \left( L_i(\omega_i^t) - L^* \right) + 4\mathcal{L}({\eta ^t})^2\tau,
\end{align}
where \(\tau  = \sum\nolimits_{i = 1}^{\left| S \right|} {\frac{1}{{\left| S \right|}}} ({L^*} - L^*_i) = {L^*} - \sum\nolimits_{i = 1}^{\left| S \right|} {\frac{1}{{\left| S \right|}}} L^*_i\).

Let \(D =  - {\gamma ^t}\sum\limits_{i = 1}^{\left| S \right|} {\frac{1}{{\left| S \right|}}} ({L_i}(\omega _i^t) - {L^*})\). To bound $D$,
\begin{align}
	\label{eq:L1_8}
	\sum\limits_{i = 1}^{\left| S \right|} {\frac{1}{{\left| S \right|}}} ({L_i}(\omega _i^t) - {L^*}) &= \sum\limits_{i = 1}^{\left| S \right|} {\frac{1}{{\left| S \right|}}} ({L_i}(\omega _i^t) - L({\omega ^t})) + \sum\limits_{i = 1}^{\left| S \right|} {\frac{1}{{\left| S \right|}}} ({L_i}({\omega ^t}) - {L^*})\notag\\
	&\ge \sum\limits_{i = 1}^{\left| S \right|} {\frac{1}{{\left| S \right|}}} \langle g^t_i,\omega ^t_i - {\omega ^t}\rangle  + (L({\omega ^t}) - {L^*})\notag\\
	&\ge  - \frac{1}{2}\sum\limits_{i = 1}^{\left| S \right|} {\frac{1}{{\left| S \right|}}} [{\eta ^t}{\left\| {g^t_i} \right\|^2} + \frac{1}{{{\eta ^t}}}{\left\| {\omega ^t_i - {\omega ^t}} \right\|^2}] + (L({\omega ^t}) - {L^*})\notag\\
	&\ge  - \sum\limits_{i = 1}^{\left| S \right|} {\frac{1}{{\left| S \right|}}} [{\eta ^t}\mathcal{L}({L_i}({\omega ^t}) - L^*_i) + \frac{1}{{2{\eta ^t}}}{\left\| {\omega ^t_i - {\omega ^t}} \right\|^2}] + (L({\omega ^t}) - {L^*}).
\end{align}
where the first inequality is based on Assumption 2, the second inequality results from AM-GM inequality, and the third inequality results from Eq.~(\ref{eq:L1_2}).

For $\eta^t \leq \frac{1}{4\mathcal{L}}$, since \(\sum\limits_{i = 1}^{\left| S \right|} {\frac{1}{{\left| S \right|}}} ({L_i}({\omega ^t}) - {L^*}) \ge 0\), $\tau \geq 0$, we have
\begin{align}
	C &= {\gamma ^t}\sum\limits_{i = 1}^{\left| S \right|} {\frac{1}{{\left| S \right|}}} [{\eta ^t}\mathcal{L}({L_i}({\omega ^t}) - L^*_i) + \frac{1}{{2{\eta ^t}}}{\left\| {\omega ^t_i - {\omega ^t}} \right\|^2}] - {\gamma ^t}(L({\omega ^t}) - {L^*}) + 4\mathcal{L}{({\eta ^t})^2}\tau\notag \\
	&= {\gamma ^t}({\eta ^t}\mathcal{L} - 1)\sum\limits_{i = 1}^{\left| S \right|} {\frac{1}{{\left| S \right|}}} ({L_i}({\omega ^t}) - L^*_i) + (4\mathcal{L}{({\eta ^t})^2} + {\gamma^t}{\eta ^t}\mathcal{L})\tau  + \frac{{{\gamma ^t}}}{{2{\eta ^t}}}\sum\limits_{i = 1}^{\left| S \right|} {\frac{1}{{\left| S \right|}}} {\left\| {\omega ^t_i - {\omega ^t}} \right\|^2}\notag\\
	&\le 6\mathcal{L}({\eta ^t})^2\tau  + \sum\limits_{i = 1}^{\left| S \right|} {\frac{1}{{\left| S \right|}}} {\left\| {\omega ^t_i - {\omega ^t}} \right\|^2}.
\end{align}

Substituting $C$ into \({A_1}\), we obtain
\begin{align}
	\label{eq:L1_9}
	{A_1}{\rm{ }} 
	&= {\left\| {{\omega ^t} - {\omega ^*} - {\eta ^t}{{\bar g}^t}} \right\|^2}\notag\\
	&\le (1 - \mu {\eta ^t}){\left\| {{\omega ^t} - {\omega ^*}} \right\|^2} + 2\sum\limits_{i = 1}^{\left| S \right|} {\frac{1}{{\left| S \right|}}} {\left\| {\omega ^t_i - {\omega ^t}} \right\|^2} + 6\mathcal{L}({\eta ^t})^2\tau.
\end{align}

By Eq.~(\ref{eq:L1_9}), taking the expectation on both sides of Eq.~(\ref{eq:L1_0}) and simplifying the inequality, we complete the proof of Lemma 1.

\textbf{Lemma 2}. According to Theorem 1, we have
\begin{equation}
	\label{eq:2_6}
	\mathbb{E}\left\| \bar g'^t - \bar g^t \right\|^2 \le \sum_{i = 1}^{|S|} \frac{2}{|S|} \left\| g_i^t \right\|^2.
\end{equation}

\textbf{Proof of Lemma 2}. 
\begin{align}
	\mathbb{E}{\left\| {{\bar g'^t} - {\bar g^t}} \right\|^2}
	&= \mathbb{E}{\left\| {\sum\limits_{i = 1}^{\left| S \right|} {\frac{1}{{\left| S \right|}}({g_i'^t} - {g_i^t)} }} \right\|^2} \\
	&= \sum\limits_{i = 1}^{\left| S \right|} {\frac{1}{\left| S \right|^2}} \mathbb{E}{\left\| {{g_i'^t} - g_i^t} \right\|^2} \\
	&\le \sum\limits_{i = 1}^{|S|} {\frac{2}{{|S|^2}}} {\left\| {g_i^t} \right\|^2}.
\end{align}

\textbf{Lemma 3}. Assume that the step size (learning rate) $\eta^t$ is decreasing by $\eta^t \leq \alpha \eta^{t+1}$ with $\alpha > 1$, it follows that
\begin{equation}
	\label{eq:2_7}
	\mathbb{E}\left( {\sum\limits_{i = 1}^{\left| S \right|} {\frac{1}{{\left| S \right|}}{{\left\| {\omega ^t} - {\omega _i^t} \right\|}^2}} } \right) \le \alpha^2{({\eta ^t})^2}{K^2}
\end{equation}

\textbf{Proof of Lemma 3}.

Since $\eta^t$ is decreasing by $\eta^t \leq \alpha \eta^{t+1}$, we have
\begin{align}
	\mathbb{E}\sum\limits_{i = 1}^{\left| S \right|} {\frac{1}{{\left| S \right|}}} {\left\| {{\omega ^t} - \omega _i^t} \right\|^2} 
	&= \mathbb{E}\sum\limits_{i = 1}^{\left| S \right|} {\frac{1}{{\left| S \right|}}} {\left\| {(\omega _i^t - {\omega ^{{t_0}}}) - ({\omega ^t} - {\omega ^{{t_0}}})} \right\|^2}\\
	&\le \mathbb{E}\sum\limits_{i = 1}^{\left| S \right|} {\frac{1}{{\left| S \right|}}} {\left\| {\omega _i^t - {\omega ^{{t_0}}}} \right\|^2}\\
	&\le \sum\limits_{i = 1}^{\left| S \right|} {\frac{1}{{\left| S \right|}}} \mathbb{E}\sum\limits_{t = {t_0}}^{t - 1} {({\eta ^t})^2} {\left\| {g'^t_i} \right\|^2}\\
	&\le \sum\limits_{i = 1}^{\left| S \right|} {\frac{1}{{\left| S \right|}}} \mathbb{E}\sum\limits_{t = {t_0}}^{t - 1} {({\eta ^{{t_0}}})^2} {K^2}\\
	&\le \alpha^2{({\eta ^t})^2}{K^2}
\end{align}

\textbf{Proof of the convergence in Post-training}.

Let \({\Delta _{t + 1}} = \mathbb{E}{\left\| {{\omega ^{t + 1}} - {\omega ^*}} \right\|^2}\). From Lemma 1, Lemma 2, and Lemma 3, it follows that
\begin{equation}
	\label{eq:thm_1}
	{\Delta _{t + 1}} \le (1 - \mu {\eta ^t}){\Delta _t} + {({\eta ^t})^2}B.
\end{equation}

where,
\begin{equation}
	B = \sum\limits_{i = 1}^{\left| S \right|} \frac{2\|g_i^t\|^2}{\| S \|^2} + 6\mathcal{L}\tau  + 2\alpha^2{K^2}.
\end{equation}

For a step size that is decreasing, \({\eta ^t} = \frac{\beta }{{t + \gamma }}\) for some \(\beta  > \frac{1}{\mu }\) and \(\gamma  > 0\) such that \({\eta ^1} \le \min \{ \frac{1}{\mu },\frac{1}{{4\mathcal{L}}}\}  = \frac{1}{{4\mathcal{L}}}\) and \({\eta ^t} \le \alpha{\eta ^{t + 1}}\) with $\alpha > 1$. We will prove \({\Delta _t} \le \frac{v}{{\gamma  + t}}\) where v = \(\max \{ \frac{{{\beta ^2}B}}{{\beta \mu  - 1}},(\gamma  + 1){\Delta _1}\} \).

We prove it by induction. Firstly, the definition of $v$ ensures that it holds for $t = 1$. Assume the conclusion holds for some $t$, it follows that
\begin{align}
	\Delta_{t+1} &\leq (1 - \eta^t \mu) \Delta_t + ({\eta ^t})^2 B \nonumber \\
	&= \left( 1 - \frac{\beta \mu}{t + \gamma} \right) \frac{v}{t + \gamma} + \frac{\beta^2 B}{(t + \gamma)^2} \nonumber \\
	&= \frac{t + \gamma - 1}{(t + \gamma)^2} v + \left[ \frac{\beta^2 B}{(t + \gamma)^2} - \frac{\beta \mu - 1}{(t + \gamma)^2} v \right] \nonumber \\
	&\leq \frac{v}{t + \gamma + 1}.
\end{align}

Then from eq.~(\ref{eq:2_7}),
\begin{equation}
	\mathbb{E}[L({\omega ^t})] - {L^*} \le \frac{\mathcal{L}}{2}{\Delta _t} \le \frac{\mathcal{L}}{2} \cdot \frac{v}{{\gamma  + t}}
\end{equation}

Specifically, if we choose \(\beta  = \frac{2}{\mu }\), \(\gamma  =  8\frac{\mathcal{L}}{\mu } - 1 \) and denote \(\kappa  = \frac{\mathcal{L}}{\mu }\), then \({\eta ^t} = \frac{2}{\mu } \cdot \frac{1}{{\gamma  + t}}\) and
\begin{equation}
	\mathbb{E}[L({\omega ^t})] - {L^*} \le \frac{{2\kappa }}{{\gamma  + t}}(\frac{B}{\mu } + 2\mathcal{L}{\Delta _1}),
\end{equation}
which implies 
\begin{equation}
	\lim_{t\to\infty} ||L(\omega^t) - L^*||= 0.
\end{equation}

Therefore, in the post-training stage, FedOSD can converge to the local optimum of Problem (\ref{FL_obj}).

\subsection{Related Work and Comparison} \label{sec:related_work}
In this section, we delineate the distinctions between our proposed FedOSD and some previous FU algorithms. For handling the issue of performing GA to unlearn in the unlearning stage, there are related works such as EWCSGA \cite{EWCSGA}, FUPGA \cite{FUPGA}, and SFU \cite{SFU}.

[EWCSGA]. EWCSGA incorporates a regularization term to the cross entropy loss to mitigate the negative impact on the model utility. Specifically, the added regularization term can lead to reducing the norm of the local gradient of $g_u^t$ to limit the model update. Hence, it can mitigate the adverse effects of the gradient explosion of GA in a way.

[FUPGA]. FUPGA projects model parameters to an $L_2$-norm ball of radius $\delta$. But it requires experimentally tuning $\delta$, and a fixed $\delta$ cannot guarantee the model convergence.

[SFU]. SFU applies the idea of incremental learning to FU by projecting the gradient of the target client to a vector that is orthogonal to the subspace formed by the remaining clients' representation matrices. Specifically, for the operation $y=Wx$ inside the Deep Learning model, SFU aims to obtain a $\Delta \omega$ that can make $(W+\Delta \omega) x - Wx = \Delta \omega$, i.e., $\Delta \omega x = 0$, so that the model update would have the same performance on the data sample $x$. To achieve this goal, it requires each remaining client to upload the output of each model layer on every selected data sample to the server, which takes huge storage and time to compute such an $\Delta \omega$. Therefore, it entails substantial computation and communication costs and raises privacy concerns. Besides, based on the formulation of SFU, it does not work in the common setting of Federated Learning where the models contain bias parameters. In conclusion, it's impractical to adopt this incremental learning method to Federated Unlearning.

\subsection{Privacy Discussion about FedOSD} \label{privacy}
FedOSD adheres to the conventional Federated Learning framework, ensuring that it does not introduce additional privacy concerns. Our privacy analysis begins with the computation of clients' local gradients. As detailed in the main paper, we follow \cite{FUPGA, FedRecovery, FedFV, FedMDFG} to employ Stochastic Gradient Descent (SGD) on clients' local data with a local epoch $E=1$. Consequently, the client $i$'s gradient $g_i^t$ is the same as the result of its local update divided by the learning rate, i.e., $g_i^t = (\omega^t - \omega_i^{t}) / \eta^t$, where $\omega_i^t$ represents the local training outcomes of client $i$ on the global model $\omega^t$. Thus, in our approach, describing the upload of $g_i^t$ to the server is effectively equivalent to uploading the local training result $\omega_i^t$, after which the server calculates $g_i^t$ by $g_i^t = (\omega^t - \omega_i^{t}) / \eta^t$. 

Note that when the local epoch $E$ exceeds 1, the value of $g_i^t = (\omega^t - \omega_i^{t}) / \eta^t$ serves as an approximation of the local gradient. We conduct additional experiments, detailed in Section ~\ref{sec:additional_exp}, to assess the unlearning effectiveness and the model utility under the setting of $E=5$.

Besides, gradient privacy protection represents a significant research direction in FL. It is generally considered safe from a privacy perspective to upload the plaintext gradient $g_i^t$ or the local training result $\omega_i^t$ to the server when the batch size exceeds 32.

Looking forward, various methods, such as Homomorphic Encryption, could be employed to encrypt $g_i^t$ (or the $\omega_i^t$) before uploading to the server, thereby enhancing the privacy safeguards of FedOSD.

\newpage

\section{Complete and Extra Experimental Results}
\subsection{Experimental Settings}
\textbf{Datasets and Models}.
In the paper, we utilize LeNet-5 \cite{MNIST} on MNIST, and adopt Multilayer perceptron (MLP) \cite{MLP} on Fashion MNIST (FMNIST \cite{fashion}), where there are three layers and each layer contains 400 neurons. For CIFAR-10, we follow \cite{FedAvg,FedFV} to implement CNN \cite{MNIST} with two convolutional layers and three fully-connected layers. Both the two convolutional layers have 64 channels, respectively, while the fully connected layers have 384, 192, and 10 neurons, respectively. For CIFAR-100 \cite{cifar}, we follow \cite{FedMDFG, FedLF} to adopt NFResNet-18 \cite{NFResnet}, which is an advanced ResNet for distributed training tasks.

\textbf{Data Partition.} We consider four types of data partitions to simulate clients under various heterogeneous scenarios.

\begin{itemize}
	\item Pat-20 \cite{FedAvg,FedFV}: Randomly assign each client the data from 20\% of the classes. For example, in a dataset like MNIST with 10 classes, each client receives data from two classes. It is possible for two clients to share data from the same class. All clients have equal amounts of data.
	\item Pat-10: It's an extreme data-island scenario where each client has 10\% of distinct classes. For example, in CIFAR-10 with 10 classes, each class's data is allocated randomly to one of 10 clients, ensuring each client has data from one distinct class. All clients have equal amounts of data.
	\item Pat-50: It constructs a scenario where each client has 50\% of the classes. Similar to other partitions, the total amount of data is evenly distributed among the clients
	\item IID: The data is randomly and equally separated among all clients.
\end{itemize}

\textbf{Baselines}. The descriptions of the baselines are shown below.
\begin{itemize}
	\item Retraining: The remaining clients adopt FedAvg \cite{FedAvg} to cooperatively train a FL global model from scratch.
	\item FedEraser \cite{FedEraser}: It is also a kind of retraining method but leverages the norms of the local updates stored in the preceding FL training to accelerate retraining. To achieve this goal, each client (or the server) should pre-store the norm of the local gradients of clients in previous FL training.
	\item FedRecovery \cite{FedRecovery}: It eases the impact of a client by removing a weighted sum of gradient residuals from the global model. To achieve this goal, each client (or the server) should pre-store the local gradient at each communication round during the previous FL training.
	\item MoDe \cite{MoDe}: An FU algorithm that adopts momentum degradation to unlearn a client.
	\item EWCSGA \cite{EWCSGA}: A GA-based FU algorithm that incorporates a regularization term to the cross entropy loss to limit the model update when unlearning.
	\item FUPGA \cite{FUPGA}: A GA-based FU algorithm that projects the model parameters to an $L_2$-norm ball of radius $\delta$ to limit the model update when unlearning.
\end{itemize}

\begin{figure}[t!]
	\centering
	\subfigure[MNIST] {
		\label{fig:a}     
		\includegraphics[width=0.22\columnwidth]{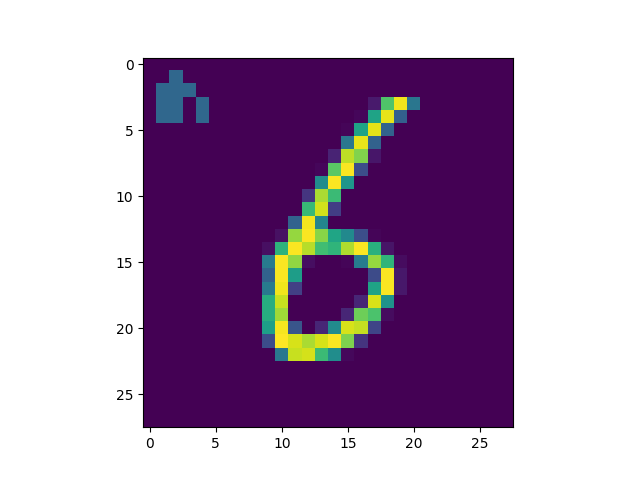}  
	}
	\subfigure[FMNIST] { 
		\label{fig:b}     
		\includegraphics[width=0.22\columnwidth]{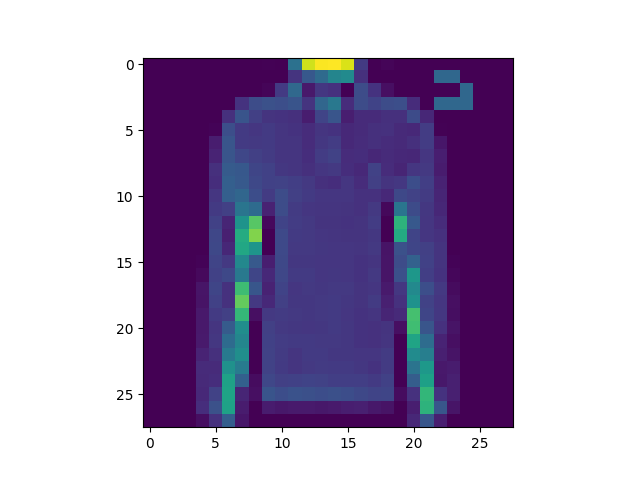}     
	}
	\subfigure[CIFAR10] {
		\label{fig:c}
		\includegraphics[width=0.22\columnwidth]{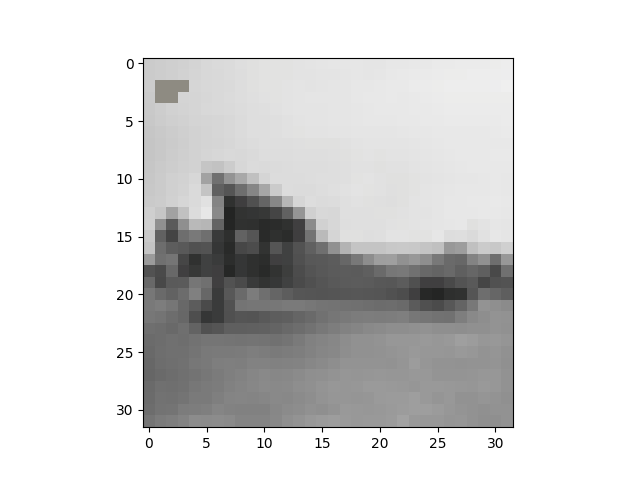}     
	}
	\subfigure[CIFAR100] {
		\label{fig:d}
		\includegraphics[width=0.22\columnwidth]{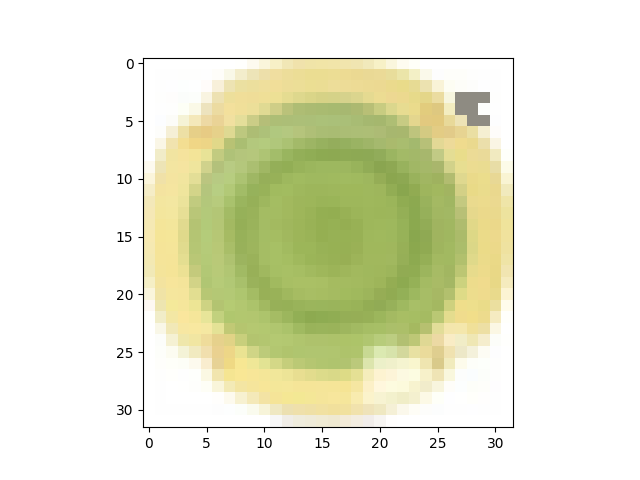}     
	}
	\caption{A demo of adding backdoor triggers on the data sample of (a) MNIST, (b) FMNIST, (c) CIFAR-10, and (d) CIFAR-100.}
	\label{fig:trigger}
\end{figure}

\textbf{Evaluation Metrics}. We adopt the model test accuracy on the retained clients (denoted as R-Acc) to evaluate the model utility. To assess the effectiveness of unlearning, we follow \cite{FUPGA,SFU,MoDe} to implant backdoor triggers into the model by poisoning the target client's training data and flipping the labels (for example, flipping Label `1' to `6'). A demo of the added trigger is presented in Fig.~\ref{fig:trigger}. As a result, the global model becomes vulnerable to the backdoor trigger. The accuracy of the model on these data measures the attack success rate (denoted as ASR), where a low ASR indicates the effective unlearning performance by the algorithm.

\textbf{Implementation Details}. We adhere to the commonly used hyper-parameters as reported in the literature for the respective algorithms. For example, the hyper-parameter $\lambda$ for MoDe is set to 0.95, following the recommendation in \cite{MoDe}. The $\delta$ of FUPGA is set to be one third of the average Euclidean distance between $\omega^0$ and a random model, with the average computed over 10 random models, as specified in \cite{FUPGA}. Consistent with the settings of \cite{FUPGA,FedRecovery}, all clients utilize Stochastic Gradient Descent (SGD) on local datasets with a local epoch $E=1$. We set the batch size as 200 and the learning rate $\eta \in \{0.005, 0.025, 0.001, 0.0005\}$ decay of 0.999 per round, where the best performance of each method is chosen in comparison. Prior to unlearning, we run FedAvg \cite{FedAvg} for 2000 communication rounds to generate the original model $\omega^0$ for unlearning. The maximum unlearning round is 100, while the maximum total communication round (including unlearning and post-training) is 200. The target unlearning client $u$ is randomly selected from ten clients. For retraining-based algorithms (Retraining and FedEraser), the communication round is set to 200, the learning rates are set to the same as the previous pertaining procedure, with results averaged over five runs using different random seeds. All experiments are implemented on a server with Intel(R) Xeon(R) Platinum 8352Y CPU and NVidia(R) A800 GPU.

\subsection{Full Experimental Results}
The full experimental results of `Table 1' from the main paper are presented here, with additional data on the worst and the best R-Acc of clients. Due to the extensive number of columns, we separate the results into Table~\ref{exp1:FMNIST} and Table~\ref{exp1:CIFAR10}. Besides, we present the experimental results on MNIST and CIFAR-100 in Table~\ref{exp1:MNIST} and Table~\ref{exp1:CIFAR100}, respectively, under the same settings. These results verify that FedOSD outperforms previous algorithms in terms of the unlearning effectiveness and the model utility protection. Moreover, from the std. and the worst and the best R-Acc results, it can be seen that FedOSD has a fairer influence on the R-Acc across the remaining clients.

\begin{figure*}[t!]
	\centering
	\includegraphics[width=1.0\linewidth]{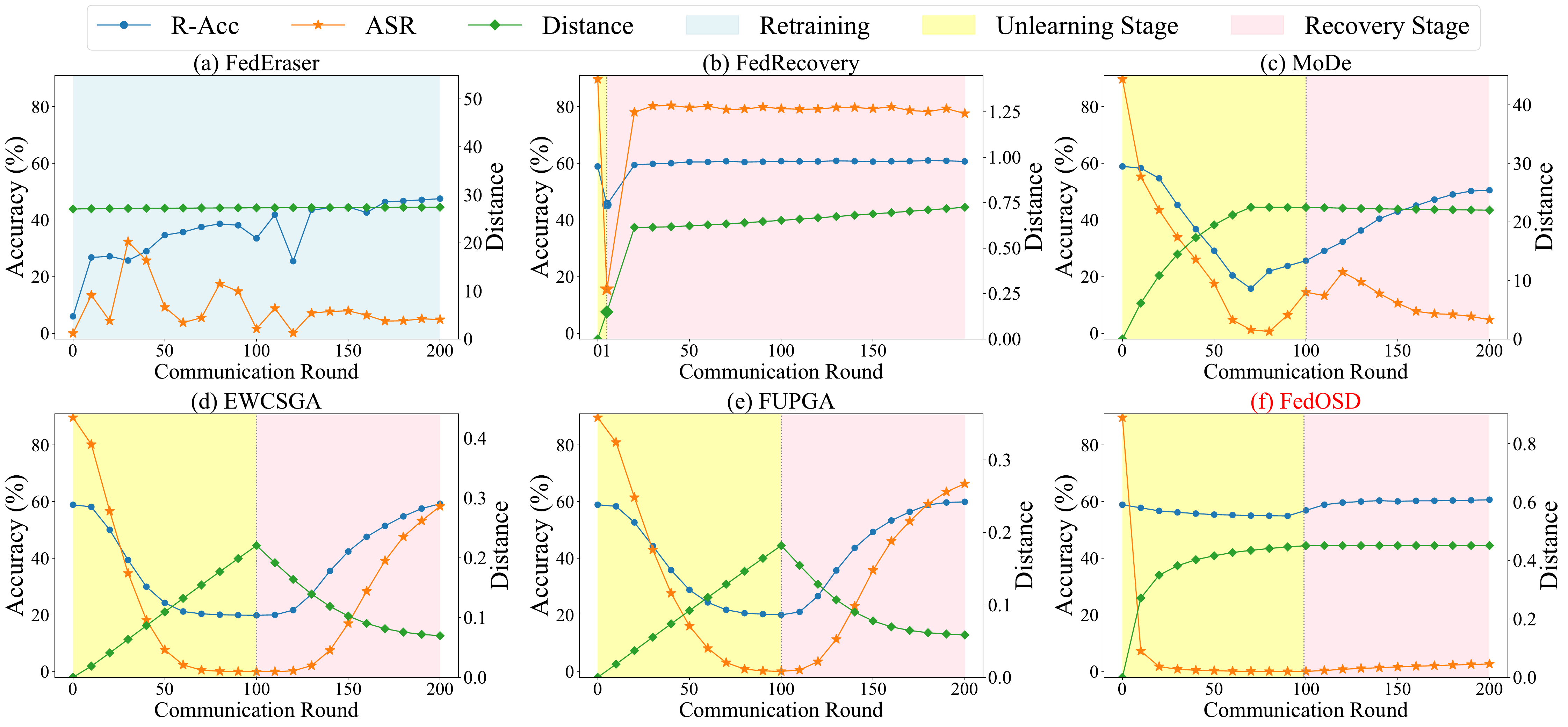}
	\caption{The ASR, mean R-Acc, and the distance away from $\omega^0$ during unlearning and post-training on CIFAR-10 Pat-20.}
	\label{fig:exp2_pat_2}
\end{figure*}

\begin{figure*}[t!]
	\centering
	\includegraphics[width=1.0\linewidth]{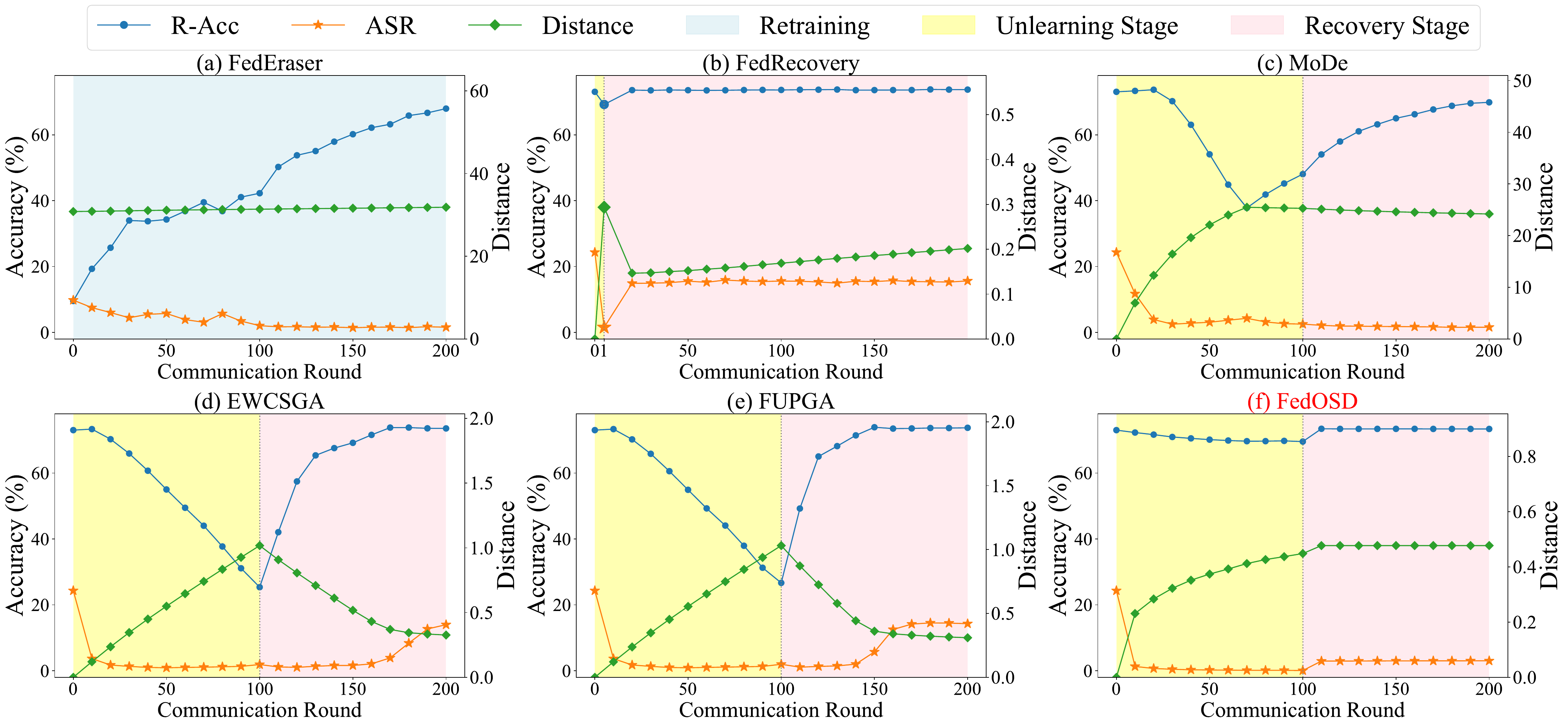}
	\caption{The ASR, the mean R-Acc, and the distance away from $\omega^0$ during unlearning and post-training on CIFAR-10 IID.}
	\label{fig:exp2_iid}
\end{figure*}

Furthermore, we supplement the visualization of the curves of the ASR, the average R-Acc, and the distance between $\omega^t$ and $\omega^0$ during both the unlearning and post-training phases on CIFAR-10 Pat-20 in Fig.~\ref{fig:exp2_pat_2} and CIFAR-10 IID in Fig.~\ref{fig:exp2_iid}. The results demonstrate that FedOSD successfully and rapidly achieves a zero ASR while maintaining the highest R-Acc during unlearning. The distance curve in the post-training stage verifies that FedRecovery, EWCSGA, and FUPGA suffer the model reverting issue, which leads to an increase in ASR during post-training, effectively negating the unlearning achievements.

Finally, we include an ablation study that omits the division by 2 in the proposed Unlearning Cross-Entropy loss (\ref{equ:UCE_Loss}). We denote this modification as M6, and list the experimental results of the unlearning in Table~\ref{appendix_tab:ablation} under the same conditions as those described in `Table 3' of the main paper. It can be seen that removing the division by 2 from the proposed UCE loss leads to the gradient explosion at the onset of the unlearning stage, which significantly deteriorates the R-Acc. The results corroborate the discussion of Section 3.1 of the main paper.

\begin{table}[t!]
	\centering
	\resizebox{1\linewidth}{!}{
		\begin{tabular}{l|l|c|c|c|l|c|c|c|l|c|c|c}
			\toprule
			& \multicolumn{4}{c|}{Pat-2}                                             & \multicolumn{4}{c|}{Pat-5}                                                                                     & \multicolumn{4}{c}{IID}                                                                   \\
			\midrule
			Algorithm    & \multicolumn{1}{l|}{ASR} & \multicolumn{1}{c|}{Utility} & Worst & Best  & \multicolumn{1}{l|}{ASR} & \multicolumn{1}{c|}{Utility} & \multicolumn{1}{l|}{Worst} & \multicolumn{1}{l|}{Best}  & \multicolumn{1}{l|}{ASR} & \multicolumn{1}{c|}{Utility} & \multicolumn{1}{l|}{Worst} & Best  \\
			\midrule
			$\omega^0$           & .991                   & .852(.113)                 & .569 & .964 & .957                   & .869(.013)                 & .844                     & .89                      & .893                   & .898(.010)                 & .878                     & .915 \\
			Retraining   & .004                   & .760(.228)                 & .375 & .948 & .002                   & .817(.025)                 & .771 & .851 & .002                   & .840(.015)                 & .812                     & .866 \\
			FedEraser    & .005                   & .763(.171)                 & .377 & .943 & .011                   & .810(.102)                 & .668                     & .918                     & .002                   & .872(.009)                 & .854                     & .884 \\
			\midrule
			FedRecovery$^1$  & .637                   & .761(.279)                 &.000    & .961 & .693                   & .823(.092)                 & .696                     & .926                     & .498                   & .871(.012)                 & .854 & .886 \\
			MoDe$^1$         & .003                   & .667(.246)                 &.000    & .878 & .005                   & .777(.046)                 & .702                     & .841                     & .002                   & .792(.012)                 & .769                     & .814 \\
			EWCSGA$^1$       &.000                      & .255(.259)                 &.000    & .729 &.000                      & .233(.261)                 &.000                        & .538                     & .101                   & .126(.009)                 & .110  & .135 \\
			FUPGA$^1$        &.000                      & .227(.254)                 &.000    & .693 &.000                      & .178(.200)                 &.000                        & .426                     & .101                   & .105(.008)                 & .092                     & .118 \\
			FedOSD$^1$       &.000                      & .757(.187)                 & .314 & .921 &.000                      & .806(.042)                 & .746 & .863                     &.000                      & .884(.011)                 & .857                     & .894 \\
			\midrule
			FedRecovery$^2$ & .960$^r$                    & .857(.112)                 & .571 & .954 & .873$^r$                   & .876(.013)                 & .860                      & .899                     & .806$^r$                   & .898(.011)                 & .877                     & .912 \\
			MoDe$^2$        & .007                   & .744(.252)                 & .382 & .951 & .003                   & .816(.028)                 & .761 & .857 & .002                   & .843(.014)                 & .817                     & .867 \\
			EWCSGA$^2$      & .935$^r$                   & .836(.173)                 & .403 & .967 & .400$^r$                     & .869(.013)                 & .844                     & .888                     & .378$^r$                   & .896(.012)                 & .870                      & .911 \\
			FUPGA$^2$       & .857$^r$                   & .837(.185)                 & .358 & .971 & .745$^r$                   & .875(.013)                 & .854                     & .896                     & .199$^r$                   & .894(.009)                 & .875 & .907 \\
			FedOSD$^2$      & .023                   & .851(.105)                 & .593 & .947 & .021                   & .874(.014)                 & .852                     & .895                     & .004                   & .897(.011)                 & .874 & .910  \\
			\bottomrule
		\end{tabular}
	}
	\caption{The ASR, the mean R-Acc (and the std.) of the model on FMNIST. The row of $\omega^0$ denotes the initial state before unlearning. The `1' marked following the algorithm name represents the results after unlearning, while `2' denotes the results after post-training. The signal `$r$' in the columns of ASR ignifies an increase of the ASR value because of the model reverting during post-training.}
	\label{exp1:FMNIST}
\end{table}

\begin{table}[t!]
	\centering
	\resizebox{1\linewidth}{!}{
		\begin{tabular}{l|l|c|c|c|l|c|c|c|l|c|c|c}
			\toprule
			& \multicolumn{4}{c|}{Pat-2}                                             & \multicolumn{4}{c|}{Pat-5}                                                                                     & \multicolumn{4}{c}{IID}                                                                   \\
			\midrule
			Algorithm    & \multicolumn{1}{l|}{ASR} & \multicolumn{1}{c|}{Utility} & Worst & Best  & \multicolumn{1}{l|}{ASR} & \multicolumn{1}{c|}{Utility} & \multicolumn{1}{l|}{Worst} & \multicolumn{1}{l|}{Best}  & \multicolumn{1}{l|}{ASR} & \multicolumn{1}{c|}{Utility} & \multicolumn{1}{l|}{Worst} & Best  \\
			\midrule
			$\omega^0$           & .897                   & .589(.115)                 & .441 & .765 & .754                   & .658(.016)                 & .629 & .683 & .243                   & .731(.013)                 & .709 & .756 \\
			Retraining   & .047                   & .507(.106)                 & .380  & .692 & .009                   & .583(.149)                 & .435 & .770  & .022                   & .511(.013)                 & .491                     & .531 \\
			FedEraser    & .098                   & .454(.158)                 & .199 & .632 & .026                   & .571(.128)                 & .399                     & .693                     & .016                   & .683(.012)                 & .661 & .696 \\
			\midrule
			FedRecovery$^1$  & .156                   & .454(.337)                 & .100   & .867 & .102                   & .476(.346)                 & .082                     & .794                     & .015                   & .692(.017)                 & .658                     & .718 \\
			MoDe$^1$         & .145                   & .256(.162)                 & .000     & .477 & .066                   & .199(.119)                 & .062                     & .321                     & .025                   & .481(.018)                 & .456 & .507 \\
			EWCSGA$^1$       & .000                       & .199(.372)                 & .000     & .897 & .000                       & .381(.426)                 & .000                         & .880                      & .018                   & .259(.010)                 & .238                     & .271 \\
			FUPGA$^1$        & .000                       & .202(.373)                 & .000     & .898 & .000                       & .388(.433)                 & .000                         & .889                     & .019                   & .271(.013)                 & .245 & .285 \\
			FedOSD$^1$       & .000                       & .549(.185)                 & .255 & .792 & .000                       & .602(.175)                 & .433                     & .803                     & .000                       & .696(.016)                 & .676                     & .717 \\
			\midrule
			FedRecovery$^2$ & .785$^r$                   & .607(.119)                 & .427 & .730  & .598$^r$                   & .643(.138)                 & .477 & .785 & .155$^r$                   & .737(.016)                 & .710                      & .763 \\
			MoDe$^2$        & .060                    & .519(.117)                 & .391 & .697 & .035                   & .582(.173)                 & .361                     & .747                     & .016                   & .703(.016)                 & .678 & .733 \\
			EWCSGA$^2$      & .581$^r$                   & .591(.194)                 & .260  & .829 & .592$^r$                   & .652(.118)                 & .497                     & .781                     & .140$^r$                    & .736(.016)                 & .712 & .763 \\
			FUPGA$^2$       & .662$^r$                   & .599(.157)                  & .331 & .809 & .602$^r$                   & .658(.091)                 & .538                     & .766                     & .144$^r$                   & .737(.014)                 & .713 & .763 \\
			FedOSD$^2$      & .027                   & .606(.101)                 & .456 & .756 & .016                   & .659(.017)                 & .627 & .689 & .030                    & .734(.015)                 & .712                     & .760  \\
			\bottomrule
		\end{tabular}
	}
	\caption{The ASR, the mean R-Acc (and the std.) of the model on CIFAR10. The row of $\omega^0$ denotes the initial state before unlearning. The `1' marked following the algorithm name represents the results after unlearning, while `2' denotes the results after post-training. The signal `$r$' in the columns of ASR ignifies an increase of the ASR value because of the model reverting during post-training.}
	\label{exp1:CIFAR10}
\end{table}

\begin{table}[t!]
	\centering
	\resizebox{1\linewidth}{!}{
		\begin{tabular}{l|l|c|c|c|l|c|c|c|l|c|c|c}
			\toprule
			& \multicolumn{4}{c|}{Pat-2}                                             & \multicolumn{4}{c|}{Pat-5}                                                                                     & \multicolumn{4}{c}{IID}                                                                   \\
			\midrule
			Algorithm    & \multicolumn{1}{l|}{ASR} & \multicolumn{1}{c|}{Utility} & Worst & Best  & \multicolumn{1}{l|}{ASR} & \multicolumn{1}{c|}{Utility} & \multicolumn{1}{l|}{Worst} & \multicolumn{1}{l|}{Best}  & \multicolumn{1}{l|}{ASR} & \multicolumn{1}{c|}{Utility} & \multicolumn{1}{l|}{Worst} & Best  \\
			\midrule
			$\omega^0$           & .997 & .963(.013) & .950  & .988 & .993 & .977(.007) & .970  & .989 & .816 & .985(.004) & .977 & .991 \\
			Retraining   & .020  & .893(.068) & .774 & .972 & .020  & .959(.009) & .945 & .973 & .005 & .981(.006) & .969 & .989 \\
			FedEraser    & .025 & .869(.061) & .812 & .980  & .047 & .944(.044)  & .891 & .992 & .006 & .979(.004) & .973 & .986 \\
			\midrule
			FedRecovery$^1$  & .038 & .716(.302) & .000     & .969 & .781 & .962(.024) & .931 & .989 & .025 & .971(.006) & .964 & .980  \\
			MoDe$^1$         & .039 & .723(.149) & .443 & .951 & .072 & .877(.053) & .797 & .942 & .016 & .944(.010) & .921 & .954 \\
			EWCSGA$^1$       & .000     & .527(.351) & .006 & .971 & .000     & 441(.485)   & .000     & .993 & .010  & .920(.009) & .907 & .938 \\
			FUPGA$^1$        & .000     & .532(.360) & .010  & .953 & .000     & .440(.491) & .000     & .993 & .009 & .946(.010) & .934 & .967 \\
			FedOSD$^1$       & .000     & .924(.048) & .808 & .973 & .000     & .927(.015) & .905 & .949 & .000     & .982(.004) & .976 & .988 \\
			\midrule
			FedRecovery$^2$ & .982$^r$ & .967(.012) & .946 & .988 & .950$^r$  & .979(.007) & .970  & .990  & .358$^r$ & .986(.004) & .977 & .992 \\
			MoDe$^2$        & .030  & .805(.157) & .523 & .961 & .033 & .942(.017) & .915 & .966 & .003 & .984(.003) & .979 & .988 \\
			EWCSGA$^2$      & .738$^r$ & .956(.019) & .926 & .983 & .970$^r$  & .978(.012) & .961 & .994 & .257$^r$ & .986(.004) & .977 & .994 \\
			FUPGA$^2$       & .589$^r$ & .947(.029) & .893 & .973 & .965$^r$ & .975(.012) & .958 & .991 & .425$^r$ & .986(.004) & .978 & .993 \\
			FedOSD$^2$      & .034 & .965(.010) & .952 & .983 & .015 & .973(.007) & .961 & .986 & .002 & .986(.003) & .977 & .993  \\
			\bottomrule
		\end{tabular}
	}
	\caption{The ASR, the mean R-Acc (and the std.) of the model on MNIST. The row of $\omega^0$ denotes the initial state before unlearning. The `1' marked following the algorithm name represents the results after unlearning, while `2' denotes the results after post-training. The signal `$r$' in the columns of ASR ignifies an increase of the ASR value because of the model reverting during post-training.}
	\label{exp1:MNIST}
\end{table}

\begin{table}[t!]
	\centering
	\resizebox{1\linewidth}{!}{
		\begin{tabular}{l|l|c|c|c|l|c|c|c|l|c|c|c}
			\toprule
			& \multicolumn{4}{c|}{Pat-2}                                             & \multicolumn{4}{c|}{Pat-5}                                                                                     & \multicolumn{4}{c}{IID}                                                                   \\
			\midrule
			Algorithm    & \multicolumn{1}{l|}{ASR} & \multicolumn{1}{c|}{Utility} & Worst & Best  & \multicolumn{1}{l|}{ASR} & \multicolumn{1}{c|}{Utility} & \multicolumn{1}{l|}{Worst} & \multicolumn{1}{l|}{Best}  & \multicolumn{1}{l|}{ASR} & \multicolumn{1}{c|}{Utility} & \multicolumn{1}{l|}{Worst} & Best  \\
			\midrule
			$\omega^0$           & .584                   & .394(.046)                 & .282 & .454 & .433 & .437(.032) & .374 & .470  & .305 & .485(.012) & .461 & .507 \\
			Retraining   & .012                   & .302(.059)                 & .147 & .360  & .011 & .320(.120) & .182 & .440  & .004 & .359(.008) & .339 & .368 \\
			FedEraser    & .016                   & .237(.047)                 & .120  & .290  & .012 & .297(.148) & .116 & .439 & .003 & .357(.009) & .336 & .369 \\
			\midrule
			FedRecovery$^1$  & .000                       & .214(.181)                 & .000     & .463 & .012 & .289(.210) & .040  & .503 & .005 & .321(.006) & .314 & .333 \\
			MoDe$^1$         & .027                   & .160(.050)                 & .038 & .229 & .009 & .209(.023) & .171 & .239 & .004 & .286(.011) & .267 & .307 \\
			EWCSGA$^1$       & .000                       & .093(.170)                 & .000     & .428 & .000     & .259(.288) & .001 & .616 & .013 & .019(.004) & .014 & .026 \\
			FUPGA$^1$        & .000                       & .090(.164)                 & .000     & .412 & .000     & .265(.289) & .000     & .620  & .012 & .017(.004) & .013 & .024 \\
			FedOSD$^1$       & .000                       & .369(.088)                 & .182 & .453 & .000     & .399(.079) & .315 & .530  & .000     & .431(.011) & .414 & .447 \\
			\midrule
			FedRecovery$^2$ & .297$^r$                   & .415(.042)                 & .315 & .472 & .179$^r$ & .458(.027) & .409 & .492 & .140$^r$  & .508(.013) & .491 & .531 \\
			MoDe$^2$        & .038                   & .366(.057)                 & .223 & .425 & .011 & .349(.026) & .306 & .378 & .004 & .377(.011) & .355 & .393 \\
			EWCSGA$^2$      & .317$^r$                   & .415(.062)                 & .307 & .482 & .228$^r$ & .456(.068) & .358 & .535 & .181$^r$ & .503(.012) & .490  & .521 \\
			FUPGA$^2$       & .322$^r$                   & .417(.050)                 & .310  & .487 & .237$^r$ & .455(.069) & .355 & .536 & .179$^r$ & .502(.011) & .488 & .523 \\
			FedOSD$^2$      & .036                   & .420(.037)                 & .331 & .476 & .025 & .458(.030) & .418 & .523 & .040  & .509(.010) & .497 & .532  \\
			\bottomrule
		\end{tabular}
	}
	\caption{The ASR, mean R-Acc (and std.) of the model on CIFAR-100. The row of $\omega^0$ denotes the initial state before unlearning. The `1' marked following the algorithm name represents the results after unlearning, while `2' denotes the results after post-training. The signal `$r$' in the columns of ASR ignifies an increase of the ASR value because of the model reverting during post-training.}
	\label{exp1:CIFAR100}
\end{table}

\begin{table}[t!]
	\centering
	\resizebox{1\linewidth}{!}{
		\begin{tabular}{l|l|c|c|c|l|c|c|c|l|c|c|c}
			\toprule
			& \multicolumn{4}{c|}{Pat-2}                                             & \multicolumn{4}{c|}{Pat-5}                                                                                     & \multicolumn{4}{c}{IID}                                                                   \\
			\midrule
			Algorithm    & \multicolumn{1}{l|}{ASR} & \multicolumn{1}{c|}{Utility} & Worst & Best  & \multicolumn{1}{l|}{ASR} & \multicolumn{1}{c|}{Utility} & \multicolumn{1}{l|}{Worst} & \multicolumn{1}{l|}{Best}  & \multicolumn{1}{l|}{ASR} & \multicolumn{1}{c|}{Utility} & \multicolumn{1}{l|}{Worst} & Best  \\
			\midrule
			$\omega^0$           & .995                   & .848(.083)                 & .680                      & .956                    & .968                   & .864(.023)                 & .832                     & .902                    & .940                    & .898(.011)                 & .876                     & .919                    \\
			Retraining   & .003                   & .753(.239)                 & .370                      & .952                    & .002                   & .861(.014)                 & .829                     & .880                     & .001                   & .896(.009)                 & .882                     & .908                    \\
			FedEraser    & .006                   & .775(.193)                 & .284                     & .945                    & .002                   & .853(.014)                 & .825                     & .874                    & .001                   & .892(.010)                 & .875                     & .907                    \\
			\midrule
			FedRecovery$^1$  & .740                    & .789(.282)                 & .000                         & .951                    & .727                   & .833(.085)                 & .719                     & .933                    & .626                   & .889(.012)                 & .868                     & .908                    \\
			MoDe$^1$         & .005                   & .685(.257)                 & .000                         & .905                    & .003                   & .825(.036)                 & .759                     & .870                     & .002                   & .867(.012)                 & .845                     & .883                    \\
			EWCSGA$^1$       & .000                       & .133(.197)                 & .000                         & .495                    & .000                       & .259(.145)                 & .098                     & .407                    & .186                   & .351(.016)                 & .323                     & .379                    \\
			FUPGA$^1$        & .000                       & .120(.199)                 & .000                         & .491                    & .000                       & .179(.145)                 & .016                     & .332                    & .104                   & .144(.010)                 & .130                      & .162                    \\
			FedOSD$^1$       & .000                       & .743(.172)                 & .394                     & .911                    & .000                       & .864(.013)                 & .840                      & .881                    & .000                       & .889(.011)                 & .870                      & .903                    \\
			\midrule
			FedRecovery$^2$ & .974$^r$                   & .855(.115)                 & .561                     & .961                    & .909$^r$                   & .868(.023)                 & .836                     & .910                     & .850$^r$                    & .898(.012)                 & .878                     & .919                    \\
			MoDe$^2$        & .004                   & .754(.237)                 & .398                     & .954                    & .002                   & .865(.009)                 & .850                      & .880                     & .002                   & .898(.010)                 & .882                     & .914                    \\
			EWCSGA$^2$      & .966$^r$                   & .839(.158)                 & .423                     & .967                    & .885$^r$                   & .870(.013)                 & .849                     & .889                    & .813$^r$                   & .898(.010)                 & .878                     & .917                    \\
			FUPGA$^2$       & .949$^r$                   & .840(.155)                 & .434                     & .966                    & .858$^r$                   & .868(.014)                 & .848                     & .892                    & .777$^r$                   & .898(.011)                 & .878                     & .918                    \\
			FedOSD$^2$      & .007                   & .840(.091)                 & .628                     & .940                     & .032                   & .864(.020)                 & .836                     & .899                    & .007                   & .894(.012)                 & .871                     & .915                     \\
			\bottomrule
		\end{tabular}
	}
	\caption{The ASR, mean R-Acc (and std.) of the model on FMNIST under local epoch $E=5$. The row of $\omega^0$ denotes the initial state before unlearning. The `1' marked following the algorithm name represents the results after unlearning, while `2' denotes the results after post-training. The signal `$r$' in the columns of ASR ignifies an increase of the ASR value because of the model reverting during post-training.}
	\label{exp:E_5}
\end{table}

\begin{table}[t!]
	\centering
	\begin{tabular}{l|l|c|l|c|l|c}
		\toprule
		\multicolumn{1}{l|}{}          & \multicolumn{2}{c|}{FMNIST}                              & \multicolumn{2}{c|}{CIFAR-10}                            & \multicolumn{2}{c}{CIFAR-100}                           \\ 
		\midrule
		\multicolumn{1}{l|}{Method} & \multicolumn{1}{l|}{ASR}  & \multicolumn{1}{l|}{R-Acc} & \multicolumn{1}{l|}{ASR}  & \multicolumn{1}{l|}{R-Acc} & \multicolumn{1}{l|}{ASR}  & \multicolumn{1}{l}{R-Acc} \\ 
		\midrule
		$\omega^0$                              & .957 & .869    & .754 & .658    & .433 & .437    \\
		\midrule
		FedOSD$^1$                          & \textbf{.000}     & \textbf{.806}     & \textbf{.000}     & \textbf{.602}    & \textbf{.000}     & \textbf{.399}    \\
		M6$^1$ & .200 & .110 & .202 & .108 & .021 & .011 \\
		\bottomrule
	\end{tabular}
	\caption{The ASR, the mean R-Acc of the model in the ablation studies in Pat-50. `1' marks the unlearning stage.}
	\label{appendix_tab:ablation}
\end{table}

\begin{table}[h!]
	\resizebox{1\linewidth}{!}{
		\begin{tabular}{lllllllll}
			\toprule
			& Retraining   & FedEraser    & FedRecovery   & MoDe         & EWCSGA       & FUPGA        & FedOSD             &             \\
			\midrule
			FMNIST    & 63.58(0.44)  & 62.77(0.57)  & 62.32(2.20)   & 65.85(0.46)  & 44.97(0.29)  & 43.20(0.31)  & \multicolumn{2}{l}{65.16(0.42)}  \\
			CIFAR-10  & 144.64(0.71) & 143.26(0.89) & 143.41(3.4)   & 149.71(0.76) & 85.37(0.48)  & 82.10(0.51)  & \multicolumn{2}{l}{148.68(0.44)} \\
			CIFAR-100 & 613.49(1.82) & 613.91(2.05) & 615.99(10.09) & 645.87(2.08) & 355.45(1.34) & 342.60(1.35) & \multicolumn{2}{l}{643.94(0.77)} \\
			\bottomrule
		\end{tabular}
	}
	\caption{The computation time (s) of clients (and the server) on FMNIST, CIFAR-10, and CIFAR-100.}
	\label{runtime}
\end{table}

\subsection{Additional Experimental Results} \label{sec:additional_exp}

\textbf{Performance under a larger local epoch}.
We further evaluate the test accuracy and fairness of algorithms under a larger local epoch $E=5$. Table~\ref{exp:E_5} presents the results. It can be seen that FedOSD outperforms previous FU algorithms in the unlearning stage by achieving an ASR of 0 while mitigating the model utility reduction. During the post-training stage, FedRecovery, EWCSGA, and FUPGA still suffer the model reverting issue and lose the unlearning achievement. In comparison, FedOSD can recover the model utility while preventing the model from moving back to $\omega^0$ to maintain the achievement of unlearning.

\subsection{Runtime}
In Table~\ref{runtime}, we report the actual computation time of clients (and the server) on FMNIST, CIFAR-10, and CIFAR-100 in Pat-50. It can be seen that it's not time-consuming for FedOSD to obtain the orthogonal steepest descent direction in the unlearning stage, and the gradient projection strategy during post-training does not bring much extra computation cost.

\clearpage

\small
\bibliography{aaai25bib}

\end{document}